%% file: main.tex
\documentclass[runningheads]{llncs}
\usepackage{graphicx}
\usepackage{amsmath,amsfonts,amssymb}
\usepackage{mathtools}
\usepackage{bm}
\usepackage{bbm}
\usepackage{tikz}
\usepackage{booktabs}
\usepackage[title]{appendix}
\usepackage[hidelinks]{hyperref}

\usetikzlibrary{calc,fit,positioning,shapes}
\usepackage[misc]{ifsym}
\newcommand{\kl}[2]{D_{\operatorname{KL}}\bigl(#1\;\|\;#2\bigr)}
\newcommand{\E}[2]{\mathbb{E}_{#1}\left[#2\right]}
\graphicspath{{figures/}}
\DeclareMathOperator*{\argmax}{argmax}
\DeclareMathOperator*{\argmin}{argmin}
\begin{document}
\title{Learning Disentangled Discrete Representations}
%
%\titlerunning{Learning Disentangled Discrete Representations}
\author{
   David~Friede\inst{1} (\Letter) \and %\orcidID{0000-1111-2222-3333} \and
   Christian~Reimers\inst{2} \and %\orcidID{1111-2222-3333-4444} \and
   Heiner~Stuckenschmidt\inst{1} \and %\orcidID{2222--3333-4444-5555} %\and
   Mathias~Niepert\inst{3,4} %\orcidID{3333-4444-5555-6666}
}
\authorrunning{D. Friede et al.}
\institute{
   University of Mannheim, Mannheim, Germany\\
   \email{\{david,heiner\}@informatik.uni-mannheim.de} \and
   Max Planck Institute for Biogeochemistry, Jena, Germany\\ 
   \email{creimers@bgc-jena.mpg.de} \and
   University of Stuttgart, Stuttgart, Germany\\
   \email{mathias.niepert@simtech.uni-stuttgart.de} \and
    NEC Laboratories Europe, Heidelberg, Germany \\
}
\toctitle{Learning Disentangled Discrete Representations}
\tocauthor{Friede,~D., Reimers,~C., Stuckenschmidt,~H., Niepert,~M.}
\maketitle              % typeset the header of the contribution
%
%
%
%%%%%%%%%%%%%%%%%%%%%%%%%%%%%%%%%%%%%%%%%%%%%%%%%%%%%%%%%%%%%%%%
%%%%%%%%%%%%%%%%%%%%%%%%%%%%%%%%%%%%%%%%%%%%%%%%%%%%%%%%%%%%%%%%
% Abstract & Keywords
%%%%%%%%%%%%%%%%%%%%%%%%%%%%%%%%%%%%%%%%%%%%%%%%%%%%%%%%%%%%%%%%
%%%%%%%%%%%%%%%%%%%%%%%%%%%%%%%%%%%%%%%%%%%%%%%%%%%%%%%%%%%%%%%%
%
\begin{abstract}
Recent successes in image generation, model-based reinforcement learning, and text-to-image generation have demonstrated the empirical advantages of discrete latent representations, although the reasons behind their benefits remain unclear. We explore the relationship between discrete latent spaces and disentangled representations
by replacing the standard Gaussian variational autoencoder (VAE) with a tailored categorical variational autoencoder.
We show that the underlying grid structure of categorical distributions mitigates the problem of rotational invariance associated with multivariate Gaussian distributions, acting as an efficient inductive prior for disentangled representations.
We provide both analytical and empirical findings that demonstrate the advantages of discrete VAEs for learning disentangled representations.
Furthermore, we introduce the first unsupervised model selection strategy that favors disentangled representations.
\end{abstract}
%
%
%
%
\input{paper}
\clearpage
\bibliographystyle{splncs04}
\bibliography{bibliography}
%
%
%
\include{appendix}
\end{document}

%% file: paper.tex
%
%
%
%%%%%%%%%%%%%%%%%%%%%%%%%%%%%%%%%%%%%%%%%%%%%%%%%%%%%%%%%%%%%%%%
%%%%%%%%%%%%%%%%%%%%%%%%%%%%%%%%%%%%%%%%%%%%%%%%%%%%%%%%%%%%%%%%
% Introduction
%%%%%%%%%%%%%%%%%%%%%%%%%%%%%%%%%%%%%%%%%%%%%%%%%%%%%%%%%%%%%%%%
%%%%%%%%%%%%%%%%%%%%%%%%%%%%%%%%%%%%%%%%%%%%%%%%%%%%%%%%%%%%%%%%
%
\section{Introduction}
\label{introduction}
%%%%%%%%%%%%%%%%%%%%%%%%%%%%%%%%
% New order
%%%%%%%%%%%%%%%%%%%%%%%%%%%%%%%%
Discrete variational autoencoders based on categorical distributions \cite{jang2017categorical,maddison2017concrete} or vector quantization \cite{van2017neural} have enabled recent success in large-scale image generation \cite{van2017neural,razavi2019generating}, model-based reinforcement learning \cite{hafner2020mastering,ozair2021vector,hafner2023mastering}, and perhaps most notably, in text-to-image generation models like Dall-E \cite{ramesh2021zero} and Stable Diffusion \cite{rombach2022high}. Prior work has argued that discrete representations are a natural fit for complex reasoning or planning \cite{jang2017categorical,ramesh2021zero,ozair2021vector} and has shown empirically that a discrete latent space yields better generalization behavior \cite{hafner2020mastering,friede2021efficient,rombach2022high}. Hafner et al. \cite{hafner2020mastering} hypothesize that the sparsity enforced by a vector of discrete latent variables could encourage generalization behavior. However, they admit that "we do not know the reason why the categorical variables are beneficial."

We focus on an extensive study of the \emph{structural impact} of discrete representations on the latent space. 
The disentanglement literature \cite{bengio2013representation,higgins2018towards,locatello2019challenging} provides a common approach to analyzing the structure of latent spaces. Disentangled representations \cite{bengio2013representation} recover the low-dimensional and independent ground-truth factors of variation of high-dimensional observations. Such representations promise interpretability \cite{higgins2018towards,adel2018discovering}, fairness \cite{locatello2019fairness,creager2019flexibly,trauble2021disentangled}, and better sample complexity for learning \cite{scholkopf2012causal,bengio2013representation,peters2017elements,van2019disentangled}. State-of-the-art unsupervised disentanglement methods enrich \emph{Gaussian} variational autoencoders \cite{kingma2013auto} with regularizers encouraging disentangling properties \cite{higgins2017beta,kumar2017variational,burgess2018understanding,kim2018disentangling,chen2018isolating}. Locatello et al.~\cite{locatello2019challenging} showed that unsupervised disentanglement without inductive priors is theoretically impossible. Thus, a recent line of work has shifted to weakly-supervised disentanglement \cite{locatello2019disentangling,shu2019weakly,locatello2020weakly,klindt2021towards}.

%
%
%
%%%%%%%%%%%%%%%%%%%%%%%%%%%%%%%%
% Figure 
%%%%%%%%%%%%%%%%%%%%%%%%%%%%%%%%
\begin{figure}[t]
\begin{center}
\centerline{
    \begin{tikzpicture}
        \node[inner sep=0pt] (0) at (0,-2.25)
        {\includegraphics[width=.14\linewidth]{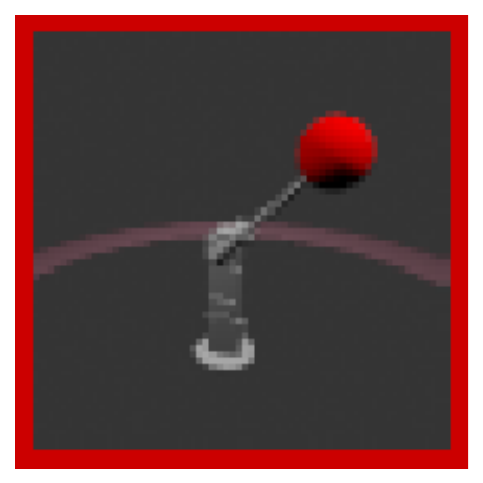}};
        \node[inner sep=0pt] (A) at (0.15,-0.2)
        {\includegraphics[width=.14\linewidth]{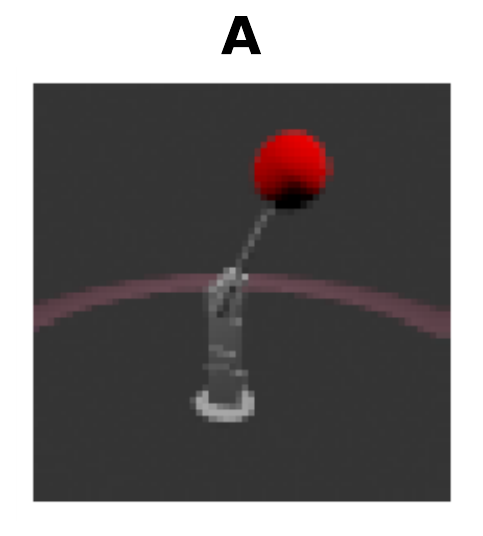}};
        \node[inner sep=0pt] (B) at (2,-2.05)
        {\includegraphics[width=.14\linewidth]{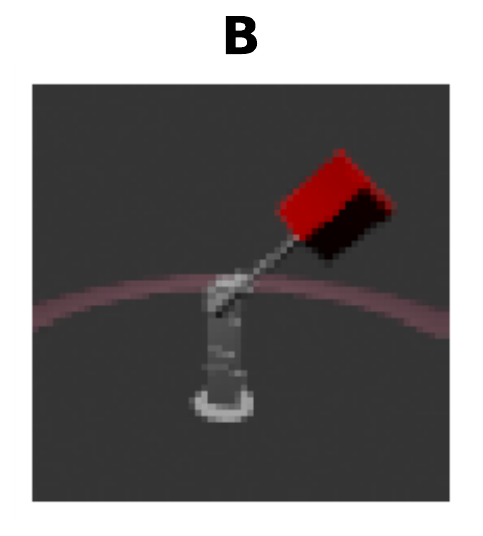}};
        \node[inner sep=0pt] (C) at (2,-0.2)
        {\includegraphics[width=.14\linewidth]{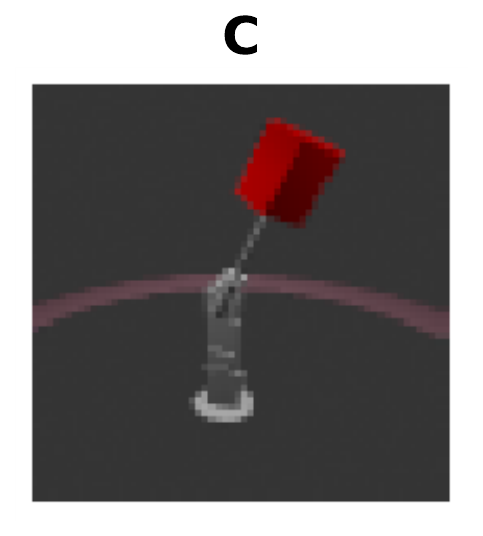}};
        \draw[->,thick] (0.north) -- (A.south);
        \draw[->,thick] (0.east) -- (B.west);
        \draw[->,thick] (0.north east) -- (C.south west);
        \node[inner sep=0pt] at (5.2,-1.3)
        {\includegraphics[width=.3\linewidth]{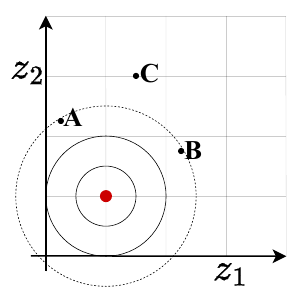}};
        \node[inner sep=0pt] at (9.1,-1.3)
        {\includegraphics[width=.3\linewidth]{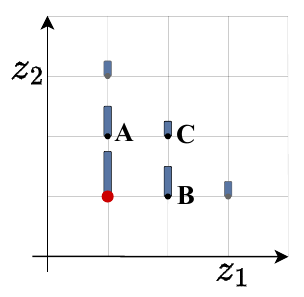}};
    \end{tikzpicture}
}
\caption{%
Four observations and their latent representation with a Gaussian and discrete VAE.
Both VAEs encourage similar inputs to be placed close to each other in latent space. 
\textbf{Left:} Four examples from the MPI3D dataset \cite{gondal2019transfer}.
The horizontal axis depicts the object's shape, and the vertical axis depicts the angle of the arm.
\textbf{Middle:} A $2$-dimensional latent space of a Gaussian VAE representing the four examples.
Distances in the Gaussian latent space are related to the Euclidean distance.
\textbf{Right:} A categorical latent space augmented with an order of the categories representing the same examples.
The grid structure of the discrete latent space makes it more robust against rotations %compared to its Gaussian counterpart, 
constituting a stronger inductive prior for disentanglement.}
\label{fig:latent_distance}%
\end{center}
\end{figure}
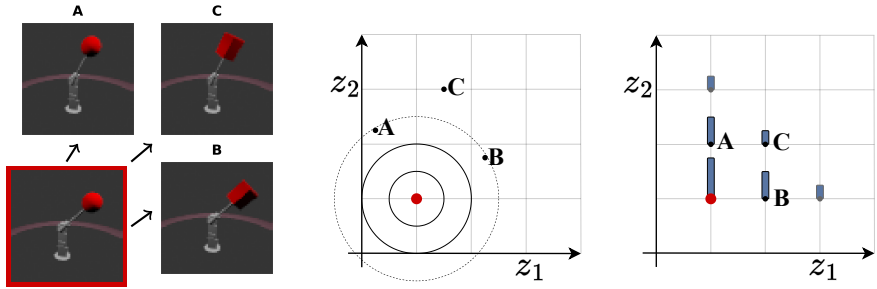
We focus on the impact on disentanglement of replacing the standard variational autoencoder with a slightly tailored \emph{categorical} variational autoencoder \cite{jang2017categorical,maddison2017concrete}.
Most disentanglement metrics assume an ordered latent space, which can be traversed and visualized by fixing all but one latent variable \cite{higgins2017beta,chen2018isolating,eastwood2018framework}. Conventional categorical variational autoencoders lack sortability since there is generally no order between the categories.
For direct comparison via established disentanglement metrics, we modify the categorical variational autoencoder to represent each category with a \emph{one-dimensional} representation.
While regularization and supervision have been discussed extensively in the disentanglement literature, the variational autoencoder is a component that has mainly remained constant.
At the same time, Watters et. al \cite{watters2019spatial} have observed that Gaussian VAEs might suffer from rotations in the latent space, which can harm disentangling properties.
We analyze the rotational invariance of multivariate Gaussian distributions in more detail and show that the underlying grid structure of categorical distributions mitigates this problem and acts as an efficient inductive prior for disentangled representations.
We first show that the observation from \cite{burgess2018understanding} still holds in the discrete case, in that neighboring points in the data space are encouraged to be also represented close together in the latent space. Second, the categorical latent space is less rotation-prone than its Gaussian counterpart and thus, constitutes a stronger inductive prior for disentanglement as illustrated in Figure~\ref{fig:latent_distance}. Third, the categorical variational autoencoder admits an unsupervised disentangling score that is correlated with several disentanglement metrics. Hence, to the best of our knowledge, we present the first disentangling model selection based on unsupervised scores.
%
%
%%%%%%%%%%%%%%%%%%%%%%%%%%%%%%%%%%%%%%%%%%%%%%%%%%%%%%%%%%%%%%%%
%%%%%%%%%%%%%%%%%%%%%%%%%%%%%%%%%%%%%%%%%%%%%%%%%%%%%%%%%%%%%%%%
% Preliminaries
%%%%%%%%%%%%%%%%%%%%%%%%%%%%%%%%%%%%%%%%%%%%%%%%%%%%%%%%%%%%%%%%
%%%%%%%%%%%%%%%%%%%%%%%%%%%%%%%%%%%%%%%%%%%%%%%%%%%%%%%%%%%%%%%%
%
\section{Disentangled Representations} \label{sct:discrete}
The disentanglement literature is usually premised on the assumption that a high-dimensional observation $\bm{x}$ from the data space $\mathcal{X}$ is generated from a low-dimensional latent variable $\bm{z}$ whose entries correspond to the dataset's ground-truth factors of variation such as position, color, or shape \cite{bengio2013representation,tschannen2018recent}. First, the \emph{independent} ground-truth factors are sampled from some distribution $\bm{z}\sim p(\bm{z})=\prod{p(z_i)}$. The observation is then a sample from the conditional probability $\bm{x}\sim p(\bm{x}|\bm{z})$. The goal of disentanglement learning is to find a representation $r(\bm{x})$ such that each ground-truth factor $z_i$ is recovered in one and only one dimension of the representation. The formalism of variational autoencoders \cite{kingma2013auto} enables an estimation of these distributions. Assuming a known prior $p(\bm{z})$, we can depict the conditional probability $p_{\theta}(\bm{x}|\bm{z})$ as a parameterized probabilistic decoder. In general, the posterior $p_{\theta}(\bm{z}|\bm{x})$ is intractable. Thus, we turn to variational inference and approximate the posterior by a parameterized probabilistic encoder $q_{\phi}(\bm{z}|\bm{x})$ and minimize the Kullback-Leibler (KL) divergence
$\kl{q_{\phi}(\bm{z}|\bm{x})}{p_{\theta}(\bm{z}|\bm{x})}$.
This term, too, is intractable but can be minimized by maximizing the evidence lower bound (ELBO)
\begin{equation}\label{eq:elbo}
    \mathcal{L}_{\theta,\phi}(\bm{x}) =
    \E{q_{\phi}(\bm{z}|\bm{x})}{\log p_{\theta}(\bm{x}|\bm{z})} -
    \kl{q_{\phi}(\bm{z}|\bm{x})}{p(\bm{z})}.
\end{equation}
State-of-the-art unsupervised disentanglement methods assume a Normal prior %distribution
$p(\bm{z}) = \mathcal{N}\bigl(\bm{0},\bm{I}\bigr)$
as well as an amortized diagonal Gaussian for the approximated posterior distribution
$q_{\phi}(\bm{z}|\bm{x})=\mathcal{N}\bigl(\bm{z}\;|\;\bm{\mu}_{\phi}(\bm{x}), \bm{\sigma}_{\phi}(\bm{x})\bm{I}\bigr)$.
They enrich the ELBO with regularizers encouraging disentangling \cite{higgins2017beta,kumar2017variational,burgess2018understanding,kim2018disentangling,chen2018isolating} and choose the representation as the mean of the approximated posterior $r(\bm{x})=\bm{\mu}_{\phi}(\bm{x})$ \cite{locatello2019challenging}.
%
%
%
%%%%%%%%%%%%%%%%%%%%%%%%%%%%%%%%
% Figure 2
%%%%%%%%%%%%%%%%%%%%%%%%%%%%%%%%
\begin{figure}[t]
\begin{center}
\centerline{
    \begin{tikzpicture}
        \node[inner sep=0pt] (0) at (0,-0.8)
        {\includegraphics[width=.4\linewidth]{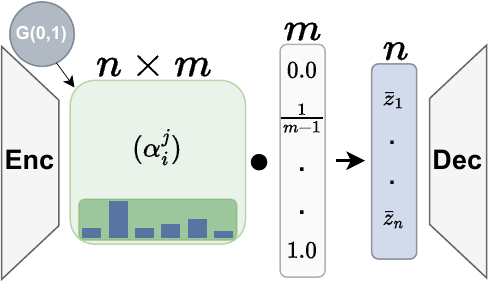}};
        \node[inner sep=0pt] (A) at (4.2,0)
        {\includegraphics[width=.18\linewidth]{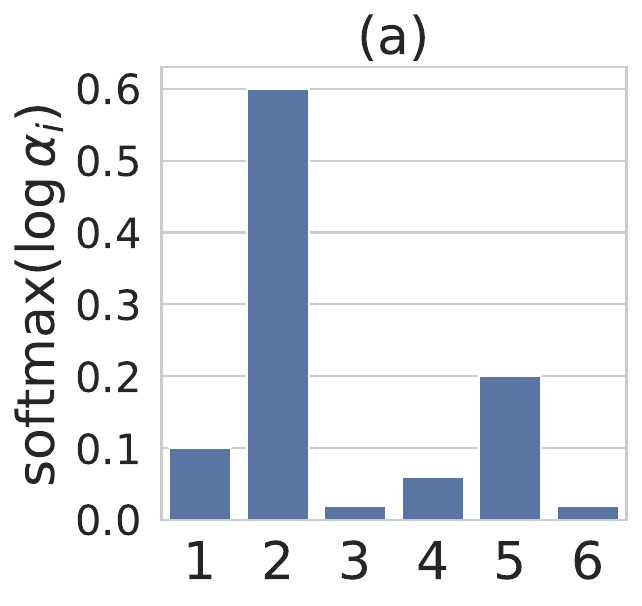}};
        \node[inner sep=0pt] (B) at (4.2,-1.95)
        {\includegraphics[width=.18\linewidth]{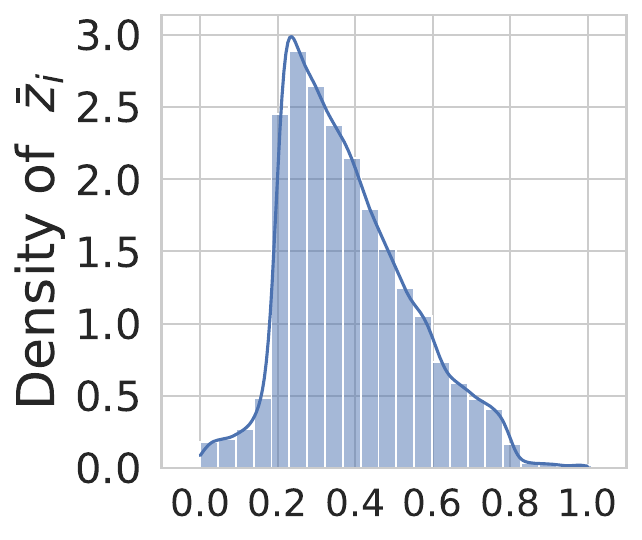}};
        \node[inner sep=0pt] (A) at (6.4,0)
        {\includegraphics[width=.18\linewidth]{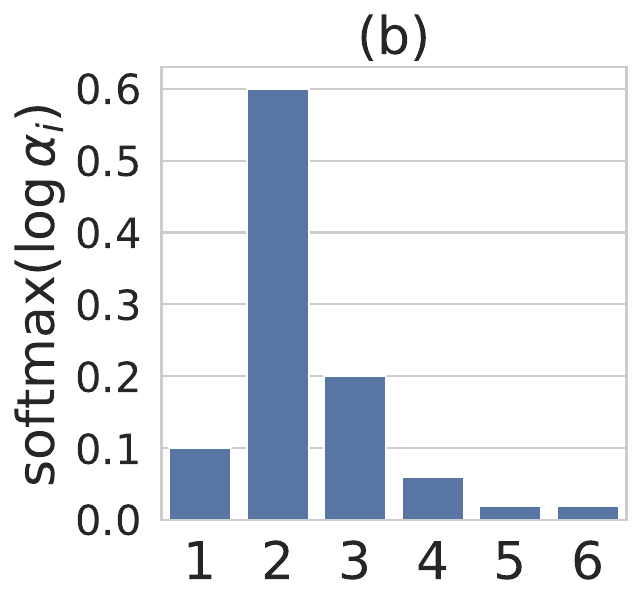}};
        \node[inner sep=0pt] (B) at (6.47,-1.95)
        {\includegraphics[width=.17\linewidth]{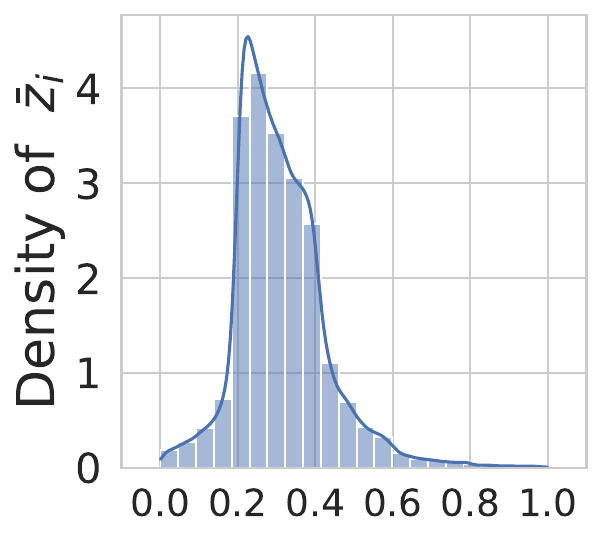}};
        \node[inner sep=0pt] (A) at (8.6,0)
        {\includegraphics[width=.18\linewidth]{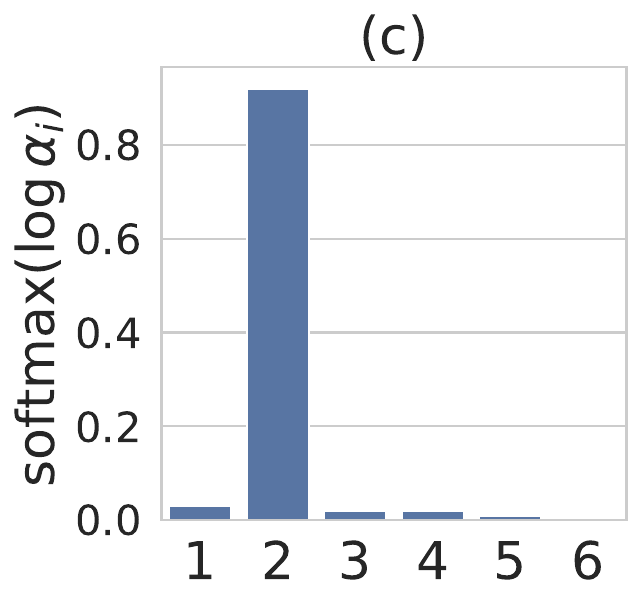}};
        \node[inner sep=0pt] (B) at (8.62,-1.95)
        {\includegraphics[width=.18\linewidth]{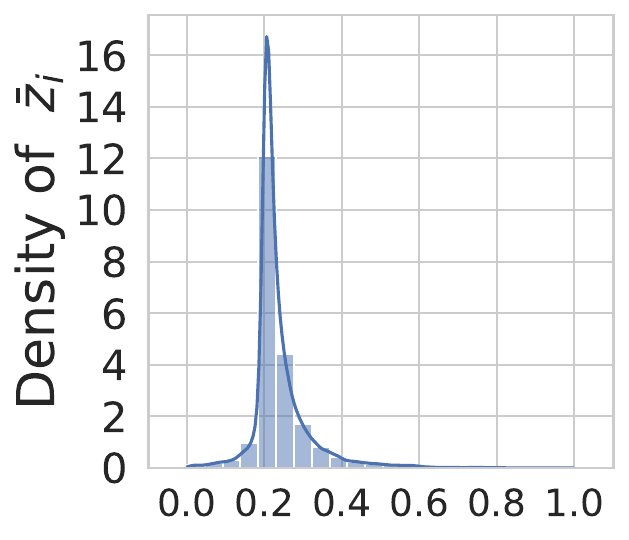}};        
    \end{tikzpicture}
}
\caption{We utilize $n$  Gumbel-softmax distributions ($\operatorname{GS}$) to approximate the posterior distribution.
\textbf{Left:} An encoder learns $nm$ parameters $a_i^j$ for the $n$ joint distributions.
Each $m$-dimensional sample %$\bm{z}_i\sim \operatorname{GS}(\bm{\alpha}_i)$ 
is mapped into the one-dimensional unit interval 
as described in Section~\ref{sec:latent_space}.
\textbf{Right:} Three examples of (normalized) parameters of a single Gumbel-softmax distribution and the corresponding one-dimensional distribution of $\bar{z}_i$.
}
\label{fig:d-vae}
\end{center}
\end{figure}
\subsubsection{Discrete VAE.}
We propose a variant of the categorical VAE modeling a joint distribution of $n$ \emph{Gumbel-Softmax} random variables \cite{jang2017categorical,maddison2017concrete}.
Let $n$ be the dimension of $\bm{z}$, $m$ be the number of categories,
$\alpha_i^j\in(0,\infty)$ be the unnormalized probabilities of the categories and $g_i^j \sim\operatorname{Gumbel}(0, 1)$ be i.i.d. samples drawn from the Gumbel distribution for $i\in[n], j\in[m]$.
For each dimension $i\in[n]$, we sample a Gumbel-softmax random variable $\bm{z}_i\sim \operatorname{GS}(\bm{\alpha}_i)$ over the simplex
$\Delta^{m-1} = \{\bm{y}\in\mathbb{R}^n\;|\;y^j\in[0,1],\sum_{j=1}^m y^j = 1\}$
by setting
\begin{equation}\label{eq:gum_softmax}
  z_i^j = \frac{\exp(\log \alpha_i^j + g_i^j)}{\sum_{k=1}^m \exp(\log \alpha_i^k + g_i^k)}
\end{equation}
for $j\in[m]$.
We set the approximated posterior distribution to be a joint distribution of $n$ Gumbel-softmax distributions, i.e., 
$q_{\phi}(\bm{z}|\bm{x})=\operatorname{GS}^n\bigl(\bm{z}\;|\;\bm{\alpha}_{\phi}(\bm{x})\bigr)$
% to be a joint distribution of $n$ Gumbel-softmax distributions
and assume a joint discrete uniform prior distribution
$p(\bm{z})=\mathcal{U}^n\{1,m\}$.
Note that $\bm{z}$ is of dimension $n\times m$.
To obtain the final $n$-dimensional latent variable $\bar{\bm{z}}$, we define a function
$f:\Delta^{m-1}\rightarrow [0,1]$
as the dot product of $\bm{z}_i$ with the vector $\bm{v}_m=(v_m^1, \dots, v_m^m)$ of $m$ equidistant entries $v_m^j = \tfrac{j-1}{m-1}$ of the interval\footnote{The choice of the unit interval is arbitrary.} $[0,1]$, i.e.,
\begin{equation}\label{eq:fn_f}
  \bar{z}_i = f(\bm{z}_i) = \bm{z}_i \cdot \bm{v}_m
  = \tfrac{1}{m-1} \textstyle\sum_{j=1}^m j z_i^j
\end{equation}
as illustrated in Figure~\ref{fig:d-vae}. We will show in Section~\ref{sct:disent_prop} that this choice of the latent variable $\bar{\bm{z}}$ has favorable disentangling properties.
The representation is obtained by the standard softmax function
$r(\bm{x})_i = f\bigl(\operatorname{softmax}(\log\bm{\alpha}_\phi(\bm{x})_i)\bigr)$.
%
%
%
%
%
%
%%%%%%%%%%%%%%%%%%%%%%%%%%%%%%%%%%%%%%%%%%%%%%%%%%%%%%%%%%%%%%%%
%%%%%%%%%%%%%%%%%%%%%%%%%%%%%%%%%%%%%%%%%%%%%%%%%%%%%%%%%%%%%%%%
% Learning disentangled discrete representations
%%%%%%%%%%%%%%%%%%%%%%%%%%%%%%%%%%%%%%%%%%%%%%%%%%%%%%%%%%%%%%%%
%%%%%%%%%%%%%%%%%%%%%%%%%%%%%%%%%%%%%%%%%%%%%%%%%%%%%%%%%%%%%%%%
%
%
%
%
\section{Learning Disentangled Discrete Representations} \label{sct:learning}
Using a discrete distribution in the latent space is a strong inductive bias for disentanglement. 
In this section, we introduce some properties of the discrete latent space and compare it to the latent space of a Gaussian VAE.
First, we show that mapping the discrete categories into a shared unit interval as in Eq.~\ref{eq:fn_f} causes an ordering of the discrete categories and, in turn, enable a definition of neighborhoods in the latent space.
Second, we derive that, %%the main argument from \cite{burgess2018understanding} still holds 
in the discrete case, neighboring points in the data space are encouraged to be represented close together in the latent space.
Third, we show that the categorical latent space is less rotation-prone than its Gaussian counterpart and thus, constituting a stronger inductive prior for disentanglement.
Finally, we describe how to select models with better disentanglement using the straight-through gap.
%
%
%
%%%%%%%%%%%%%%%%%%%%%%%%%%%%%%%%
% Neighborhoods in the data space
%%%%%%%%%%%%%%%%%%%%%%%%%%%%%%%%
%
% Moved to Appendix (app:subs:neighX)
%
%
%
%%%%%%%%%%%%%%%%%%%%%%%%%%%%%%%%
% Neighborhoods in the latent space
%%%%%%%%%%%%%%%%%%%%%%%%%%%%%%%%
%
\subsection{Neighborhoods in the latent space}\label{sec:latent_space}
In the Gaussian case, neighboring points in the observable space correspond to neighboring points in the latent space.
The ELBO Loss Eq.~\ref{eq:elbo}, more precisely the reconstruction loss as part of the ELBO, implies a topology of the observable space.
For more details on this topology, see Appendix~\ref{app:subs:neighX}.
In the case, where the approximated posterior distribution, $q_{\phi}(\bm{z}|\bm{x})$, is Gaussian and the covariance matrix, $\Sigma(\bm{x})$, is diagonal, the topology of the latent space can be defined in a similar way:
The negative log-probability is the weighted Euclidean distance to the mean $\bm{\mu}(\bm{x})$ of the distribution 
\begin{equation}
    \begin{split}
        C -\log q_{\phi}(\bm{z}|\bm{x})  
        &= \frac{1}{2}\left[ (\bm{z}-\bm{\mu}(\bm{x}))^{\intercal} \bm{\Sigma}(\bm{x})  (\bm{z}-\bm{\mu}(\bm{x}))\right]^2
        = \sum_{i=1}^n \frac{(z_i - \mu_i(\bm{x}))^2}{2\sigma_i(\bm{x})}
    \end{split}
\end{equation}
where $C$ denotes the logarithm of the normalization factor in the Gaussian density function. 
Neighboring points in the observable space will be mapped to neighboring points in the latent space to reduce the log-likelihood cost of sampling in the latent space \cite{burgess2018understanding}.

In the case of categorical latent distributions, the induced topology is not related to the euclidean distance and, hence, it does not encourage that points that are close in the observable space will be mapped to points that are close in the latent space.
The problem becomes explicit if we consider a single categorical distribution.
In the latent space, neighbourhoods entirely depend on the shared representation of the $m$ classes.
The canonical representation maps a class $j$ into the one-hot vector $\bm{e}^j = (e_1,e_2,\dots,e_m)$ with $e_k=1$ for $k=j$ and $e_k=0$ otherwise.
The representation space consists of the $m$-dimensional units vectors, and all classes have the same pairwise distance between each other.

To overcome this problem, we inherit the canonical order of $\mathbb{R}$ by depicting a $1$-dimensional representation space.
We consider the representation $\bar{z}_i=f(\bm{z}_i)$ from Eq.~\ref{eq:fn_f} that maps a class $j$ on the value $\frac{j-1}{m-1}$ inside the unit interval.
In this way, we create an ordering on the classes $1 < 2 < \dots < m$ and define the distance between two classes by $d(j,k) = \frac{1}{m-1}\vert j - k \vert$.
In the following, we discuss properties of a VAE using this representation space.
%
%
%
%%%%%%%%%%%%%%%%%%%%%%%%%%%%%%%%
% Latent space properties
%%%%%%%%%%%%%%%%%%%%%%%%%%%%%%%%
%
\subsection{Disentangling properties of the discrete VAE} \label{sct:disent_prop}
In this section, we show that neighboring points in the observable space are represented close together in the latent space and that each data point is represented discretely by a single category $j$ for each dimension $i\in\{1,\dots,n\}$.
First, we show that reconstructing under the latent variable $\bar{z}_i=f(\bm{z}_i)$ encourages each data point to utilize neighboring categories rather than categories with a larger distance.
Second, we discuss how the Gumbel-softmax distribution is encouraged to approximate the discrete categorical distribution.
For the Gaussian case, this property was shown by \cite{burgess2018understanding}.
Here, the ELBO (Eq.~\ref{eq:elbo}) depicts an inductive prior that encourages disentanglement by encouraging neighboring points in the data space to be represented close together in the latent space \cite{burgess2018understanding}.
To show these properties for the D-VAE, we use the following proposition. The proof can be found in Appendix~\ref{app:proof}.
%
%
%
%%%%%%%%%%%%%%%%%%%%%%%%%%%%%%%%
% Proposition 1
%%%%%%%%%%%%%%%%%%%%%%%%%%%%%%%%
%
\begin{proposition}\label{prop:fn_f}
  Let $\bm{\alpha}_i \in [0, \infty)^m$, $\bm{z}_i\sim \operatorname{GS}(\bm{\alpha}_i)$ be as in Eq.~\ref{eq:gum_softmax} and $\bar{z}_i=f(\bm{z}_i)$ be as in Eq.~\ref{eq:fn_f}.
  Define $j_{\text{min}}=\argmin_j\{\alpha_i^j > 0\}$ and $j_{\text{max}}=\argmax_j\{\alpha_i^j > 0\}$. Then it holds that
  \begin{enumerate}
      \item[(a)]
      $\operatorname{supp}(f)=(\frac{j_{\text{min}}}{m-1}, \tfrac{j_{\text{max}}}{m-1})$ \label{prop:fn_f_a}
      \item[(b)] $\frac{\alpha_i^j}{\sum_{k=1}^m \alpha_i^k} \rightarrow 1 \Rightarrow
      \mathbb{P}(z_i^j=1)=1 \land f(\bm{z}_i) = \mathbbm{1}_{\{\frac{j}{m-1}\}}$. \label{prop:fn_f_b} %
  \end{enumerate}%
\end{proposition}%
Prop.~\ref{prop:fn_f} has multiple consequences.
First, a class $j$ might have a high density regarding $\bar{z}_i=f(\bm{z}_i)$ although $\alpha_i^j \approx 0$. For example, if $j$ is positioned between two other classes with large $\alpha_i^k$ $\bigl($e.g. $j = 3$ in Figure~\ref{fig:d-vae}(a)$\bigr)$
Second, if there is a class $j$ such that $\alpha_i^k \approx 0$ for all $k \geq j$ or $k \leq j$, then the density of these classes is also almost zero $\bigl($Figure~\ref{fig:d-vae}(a-c)$\bigr)$.
Note that a small support benefits a small reconstruction loss since it reduces the probability of sampling a wrong class.
The probabilities of Figure~\ref{fig:d-vae} (a) and (b) are the same with the only exception that $\alpha_i^3 \leftrightarrow \alpha_i^5$ are swapped.
Since the probability distribution in (b) yields a smaller support and consequently a smaller reconstruction loss while the KL divergence is the same for both probabilities,\footnote{The KL divergence is invariant under permutation.} the model is encouraged to utilize probability (b) over (a).
This encourages the representation of similar inputs in neighboring classes rather than classes with a larger distance.

Consequently, we can apply the same argument as in \cite{burgess2018understanding}~Section~4.2 about the connection of the posterior overlap with minimizing the ELBO.
Since the posterior overlap is highest between neighboring classes, confusions caused by sampling are more likely in neighboring classes than those with a larger distance.
To minimize the penalization of the reconstruction loss caused by these confusions, neighboring points in the data space are encouraged to be represented close together in the latent space.
Similar to the Gaussian case \cite{burgess2018understanding}, we observe an increase in the KL divergence loss during training while the reconstruction loss continually decreases.
The probability of sampling confusion and, therefore, the posterior overlap must be reduced as much as possible to reduce the reconstruction loss.
Thus, later in training, data points are encouraged to utilize exactly one category while accepting some penalization in the form of KL loss,
meaning that
$\alpha_i^j/(\sum_{k=1}^m \alpha_i^k) \rightarrow 1$.
Consequently, the Gumbel-softmax distribution approximates the discrete categorical distribution, see Prop.~\ref{prop:fn_f} (b). 
An example is shown in Figure~\ref{fig:d-vae}(c).
This training behavior results in the unique situation in which the latent space approximates a discrete representation while its classes maintain the discussed order and the property of having neighborhoods.
%
%
%
%%%%%%%%%%%%%%%%%%%%%%%%%%%%%%%%
% Advantages of discrete disentanglement
%%%%%%%%%%%%%%%%%%%%%%%%%%%%%%%%
%
\subsection{Structural advantages of the discrete VAE}\label{sct:discrete_disent}
In this section, we demonstrate that the properties discussed in Section~\ref{sct:disent_prop} aid disentanglement.
So far, we have only considered a single factor $\bm{z}_i$ of the approximated posterior $q_{\phi}(\bm{z}|\bm{x})$.
To understand the disentangling properties regarding the full latent variable $\bm{z}$, we first highlight the differences between the continuous and the discrete approach.

In the continuous case, neighboring points in the observable space are represented close together in the latent space. 
However, this does not imply disentanglement, since the first property is invariant under rotations over $\mathbb{R}^n$ while disentanglement is not.
Even when utilizing a diagonal covariance matrix for the approximated posterior $q(\bm{z}|\bm{x})=\mathcal{N}\bigl(\bm{z}\;|\;\bm{\mu}(\bm{x}), \bm{\sigma}(\bm{x})\bm{I}\bigr)$, which, in general, is not invariant under rotation,
there are cases where rotations are problematic, as the following proposition shows. We provide the proof in Appendix~\ref{app:proof}.
%
%
%
%%%%%%%%%%%%%%%%%%%%%%%%%%%%%%%%
% Figure 3
%%%%%%%%%%%%%%%%%%%%%%%%%%%%%%%%
\begin{figure}[t]
\begin{center}
\centerline{
    \includegraphics[width=.17\linewidth]{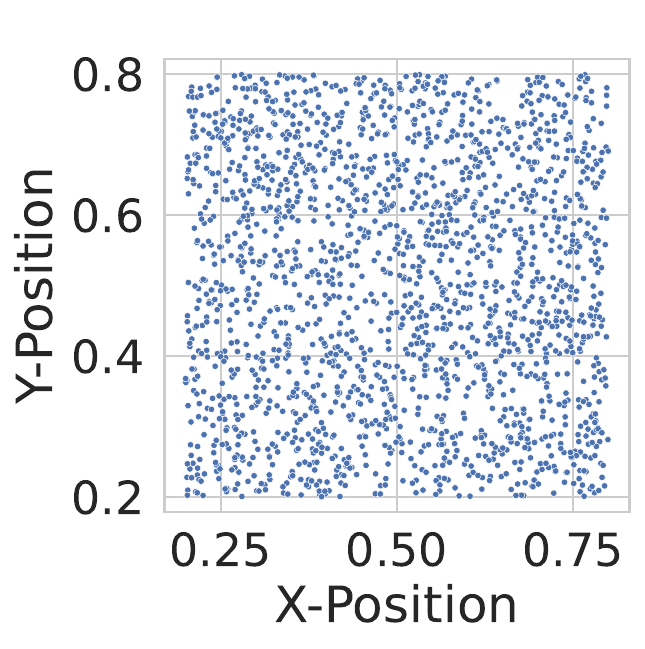}
    \hspace*{.5em} \unskip\ \vrule\ \hspace*{.5em}
    \raisebox{\dimexpr 1.3cm-0.5\height}{\rotatebox[origin=c]{90}{VAE}}\hspace{.4em}
    \includegraphics[width=.17\linewidth]{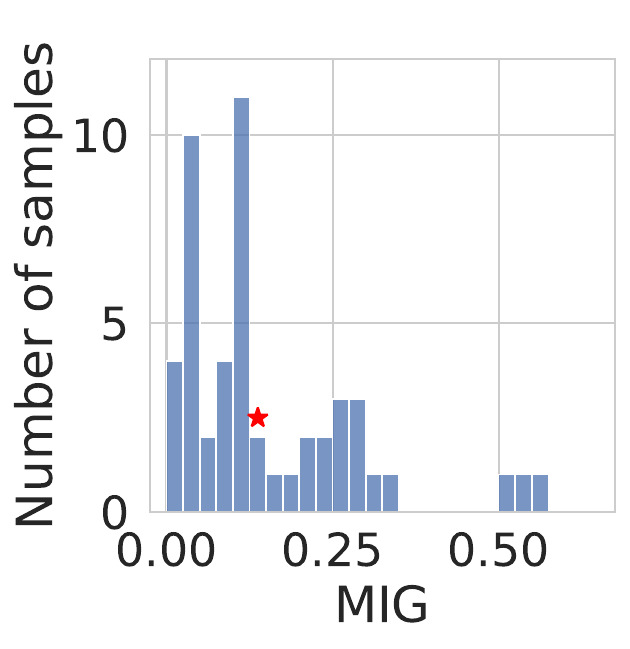}
    \includegraphics[width=.175\linewidth]{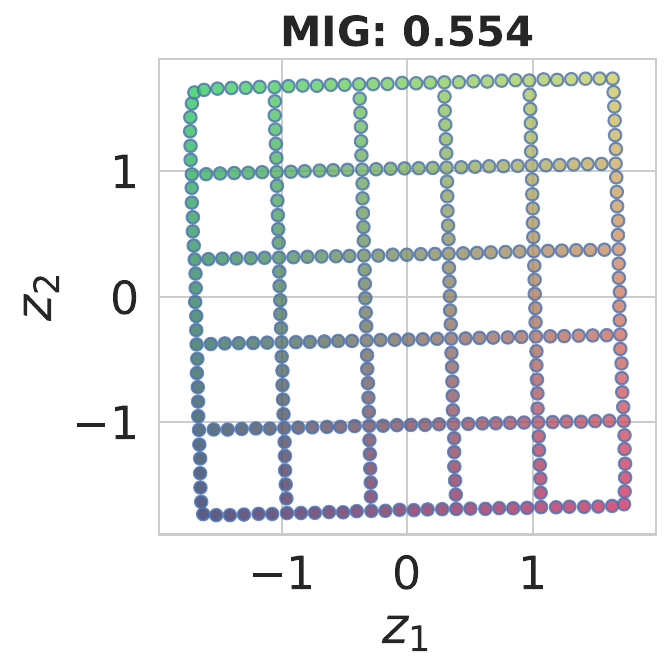}
    \includegraphics[width=.175\linewidth]{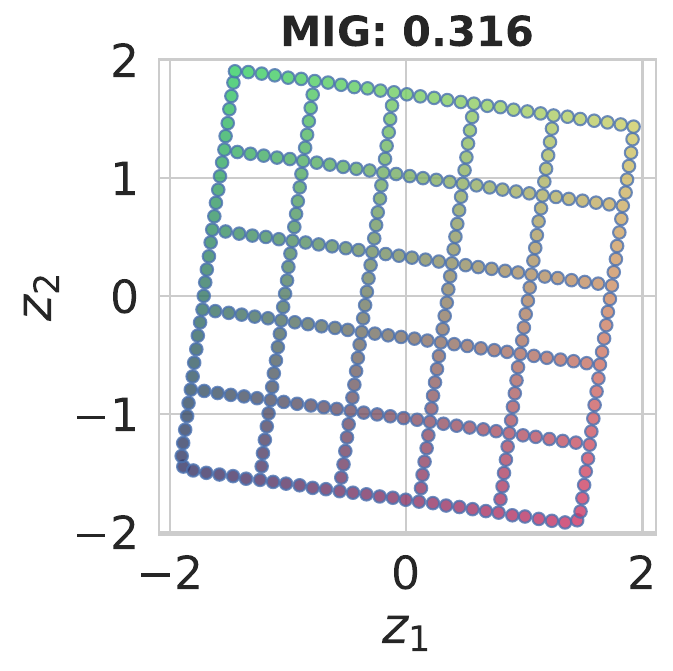}
    \includegraphics[width=.175\linewidth]{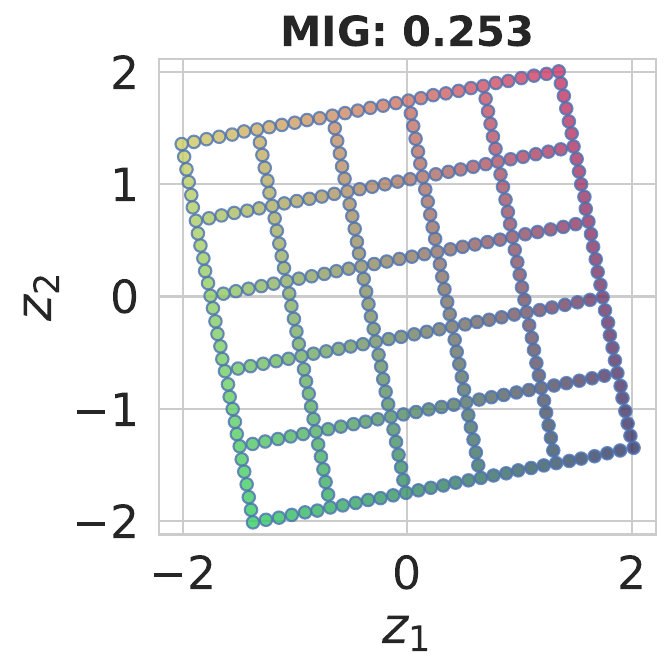}
    }
\vskip -0.015in
\centerline{
    \includegraphics[width=.17\linewidth]{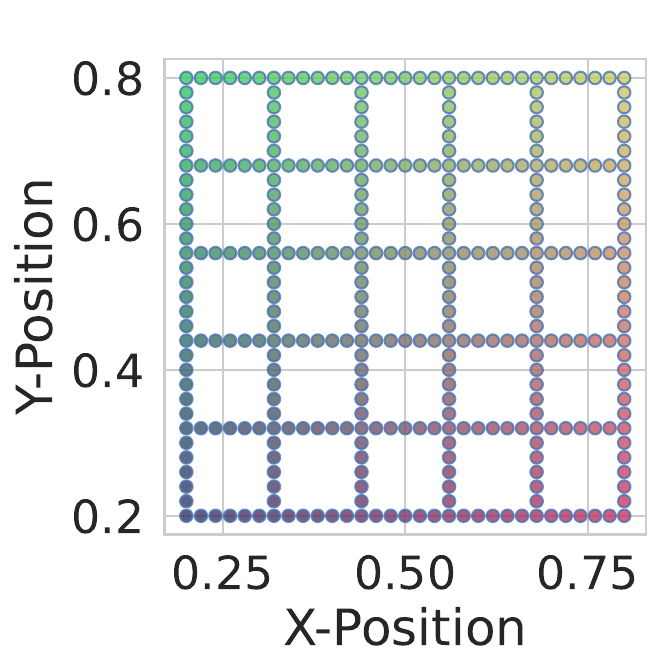}
    \hspace*{.5em} \unskip\ \vrule\ \hspace*{.5em}
    \raisebox{\dimexpr 1.35cm-0.5\height}{\rotatebox[origin=c]{90}{D-VAE}}\hspace{.4em}
    \includegraphics[width=.17\linewidth]{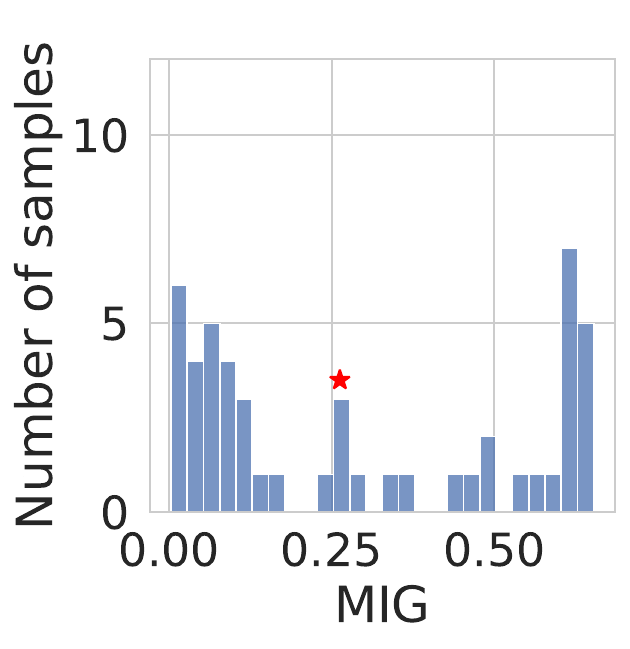}
    \includegraphics[width=.175\linewidth]{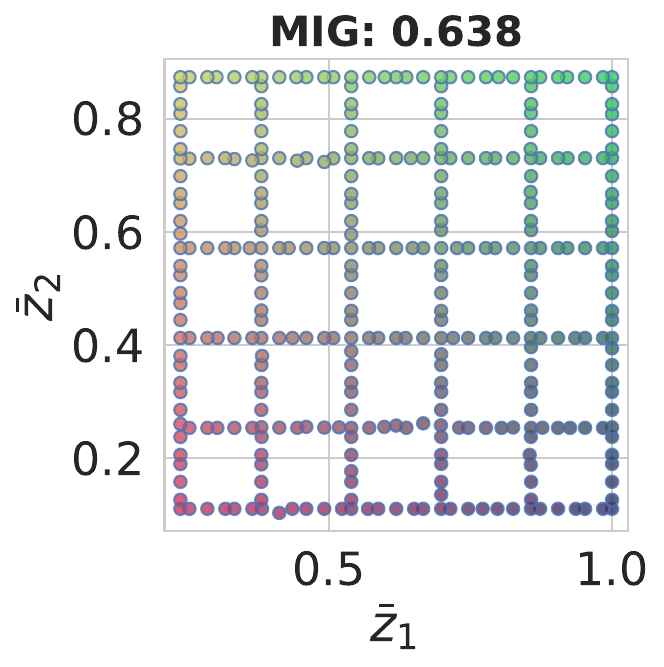}
    \includegraphics[width=.175\linewidth]{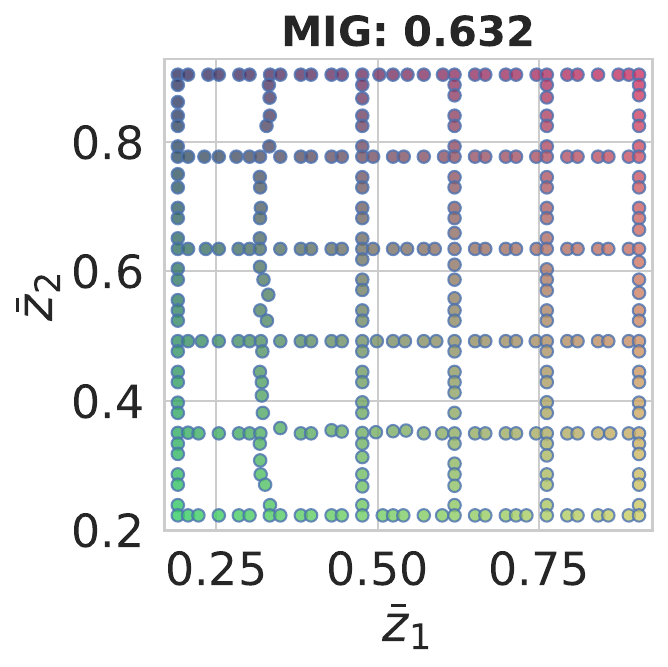}
    \includegraphics[width=.175\linewidth]{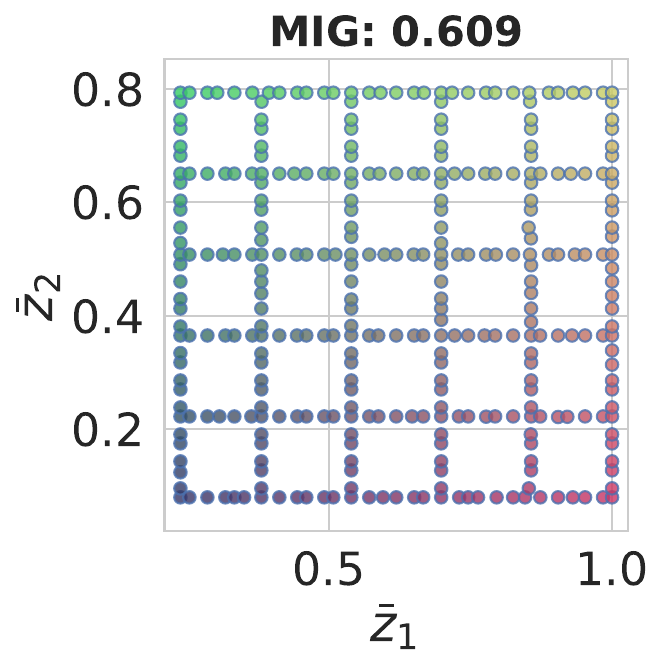}
    }
\caption{Geometry analysis of the latent space of the circles experiment \cite{watters2019spatial}.
\textbf{Col 1, top:} The generative factor distribution of the circles dataset. %is uniform in the X- and Y-positions.
\textbf{Bottom:} A selective grid of points in generative factor space spanning the data distribution.
\textbf{Col 2:} The Mutual Information Gap (MIG) \cite{chen2018isolating} for $50$ Gaussian VAE (top) and a categorical VAE (bottom), respectively.
The red star denotes the median value.
\textbf{Col 3 - 5:} The latent space visualized by the representations of the selective grid of points.
We show the best, $5$th best, and $10$th model determined by the MIG score of the Gaussian VAE (top) and the categorical VAE (bottom), respectively.}
\label{fig:circles}
\end{center}
\end{figure}
%
%
%
%
%
%
%%%%%%%%%%%%%%%%%%%%%%%%%%%%%%%%
% Proposition 2
%%%%%%%%%%%%%%%%%%%%%%%%%%%%%%%%
%
\begin{proposition}[Rotational Equivariance] \label{prop:rotation}
    Let $\alpha\in[0,2\pi)$ and let $\bm{z}\sim\mathcal{N}\bigl(\bm{\mu}, \Sigma\bigr)$ with $\Sigma=\bm{\sigma}\bm{I},\; \bm{\sigma}=(\sigma_0, \dots, \sigma_n)$.
    If $\sigma_i = \sigma_j$ for some $i\neq j\in[n]$,
    then $\bm{z}$ is equivariant under any $i,j$-rotation, i.e.,
    $R_{ij}^\alpha\bm{z} \overset{d}{=} \bm{y}$ with
    $\bm{y}\sim\mathcal{N}\bigl(R_{ij}^\alpha\bm{\mu}, \Sigma\bigr)$.%
\end{proposition}%
Since, in the Gaussian VAE, the KL-divergence term in Eq.~\ref{eq:elbo} is invariant under rotations, Prop.~\ref{prop:rotation} implies that its latent space can be arbitrarily rotated in dimensions $i,j$ that hold equal variances $\sigma_i = \sigma_j$.
Equal variances can occur, for example, when different factors exert a similar influence on the data space, e.g., X-position and Y-position or
for factors where high log-likelihood costs of potential confusion causes lead to variances close to zero.
In contrast, the discrete latent space is invariant only under rotations that are axially aligned.

We illustrate this with an example in Figure~\ref{fig:circles}. 
Here we illustrate the $2$-dimensional latent space of a Gaussian VAE model trained on a dataset generated from the two ground-truth factors, X-position and Y-position.
We train $50$ copies of the model and depicted the best, the $5$th best, and the $10$th best latent space regarding 
the Mutual Information Gap (MIG) \cite{chen2018isolating}.
All three latent spaces exhibit rotation, while the disentanglement score is strongly correlated with the angle of the rotation.
In the discrete case, the latent space is, according to Prop.~\ref{prop:fn_f} (b), a subset of the regular grid
$\mathbb{G}^n$ with $\mathbb{G}=\{\tfrac{j}{m-1}\}_{j=0}^{m-1}$
as illustrated in Figure~\ref{fig:latent_distance} (right).
Distances and rotations exhibit different geometric properties on $\mathbb{G}^n$ than on $\mathbb{R}^n$.
First, the closest neighbors are axially aligned.
Non-aligned points have a distance at least $\sqrt{2}$ times larger.
Consequently, representing neighboring points in the data space close together in the latent space encourages disentanglement.
Secondly, $\mathbb{G}^n$ is invariant only under exactly those rotations that are axially aligned.
Figure~\ref{fig:circles} (bottom right) illustrates the $2$-dimensional latent space of a D-VAE model trained on the same dataset and with the same random seeds as the Gaussian VAE model.
Contrary to the Gaussian latent spaces, the discrete latent spaces are sensible of the axes and generally yield better disentanglement scores.
The set of all $100$ latent spaces is available in Figures \ref{fig:50_vae} and \ref{fig:50_d-vae} in Appendix~\ref{app:exp}.
%
%
%%%%%%%%%%%%%%%%%%%%%%%%%%%%%%%%
% Model selection by an unspervised score
%%%%%%%%%%%%%%%%%%%%%%%%%%%%%%%%
%
\subsection{The straight-through gap} \label{sec:gap_st}
\begin{figure}[t]
\begin{center}
\centerline{
    \includegraphics[width=0.49\columnwidth]{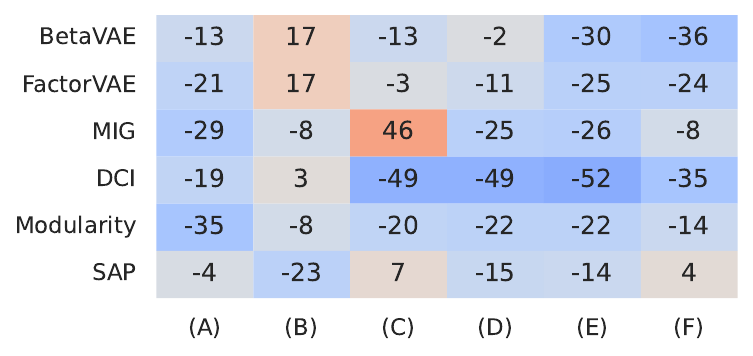}
    \includegraphics[width=0.49\columnwidth]{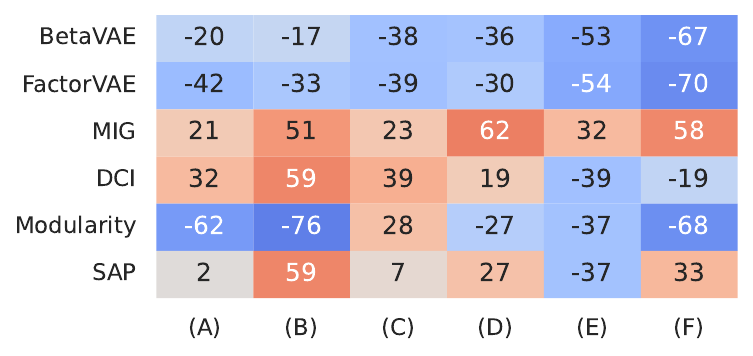}
}
\caption{The Spearman rank correlation between various disentanglment metrics and $\operatorname{Gap}_{ST}$ (\textbf{left}) and the statistical sample efficiency, i.e.,
the downstream task accuracy based on $100$ samples divided by the one on $10\,000$ samples
(\textbf{right}) on different datasets: dSprites (A), C-dSprites (B), SmallNORB (C), Cars3D (D), Shapes3D (E), MPI3D (F).
\textbf{Left:} Correlation to $\operatorname{Gap}_{ST}$ indicates the disentanglement skill.
\textbf{Right:} Only a high MIG score reliably leads to a higher sample efficiency over all six datasets.
}
\label{fig:st-gap_downstream}
\end{center}
\end{figure}

We have observed that sometimes the models approach local minima, for which $\bm{z}$ is not entirely discrete.
As per the previous discussion, those models have inferior disentangling properties.
We leverage this property by selecting models that yield discrete latent spaces.
Similar to the Straight-Through Estimator \cite{bengio2013estimating}, we round $\bm{z}$ off using $\argmax$ and measure the difference between the rounded and original ELBO, i.e.,
  $\operatorname{Gap}_{ST}(\bm{x}) = \lvert\mathcal{L}_{\theta,\phi}^{ST}(\bm{x}) - \mathcal{L}_{\theta,\phi}(\bm{x})\rvert$,
which equals zero if $\bm{z}$ is discrete.
Figure~\ref{fig:st-gap_downstream} (left) illustrates the Spearman rank correlation between $\operatorname{Gap}_{ST}$ and various disentangling metrics on different datasets.
A smaller $\operatorname{Gap}_{ST}$ value indicates high disentangling scores for most datasets and metrics.
%
%
%
%%%%%%%%%%%%%%%%%%%%%%%%%%%%%%%%
% MIG Comparison Table
%%%%%%%%%%%%%%%%%%%%%%%%%%%%%%%%
%
\begin{table*}[t]
\setlength{\tabcolsep}{3.5pt}
\caption{%
The median MIG scores in \% for state-of-the-art unsupervised methods compared to the discrete methods.
Results taken from \cite{locatello2019challenging} are marked with an asterisk~(*).
We have re-implemented all other results with the same architecture as in \cite{locatello2019challenging} for the sake of fairness.
The last row depicts the scores of the models selected by the smallest $\operatorname{Gap}_{ST}$.
The $25\%$ and the $75\%$ quantiles can be found in Table~\ref{tb:mig_25} in Appendix~\ref{app:exp}.}
\label{tb:mig}
\begin{center}
\begin{tabular}{lcccccc}
\toprule
Model & dSprites & C-dSprites & SmallNORB & Cars3D & Shapes3D & MPI3D\\
\midrule
$\beta$-VAE \cite{higgins2017beta}    &11.3$^*$& 12.5$^*$& 20.2$^*$& 9.5$^*$& n.a.& n.a.\\
$\beta$-TCVAE \cite{chen2018isolating}  &17.6$^*$& 14.6$^*$& 21.5$^*$& 12.0$^*$& n.a.& n.a.\\
DIP-VAE-I \cite{kumar2017variational}      &3.6$^*$& 4.7$^*$& 16.7$^*$& 5.3$^*$& n.a.& n.a.         \\
DIP-VAE-II \cite{kumar2017variational}     &6.2$^*$& 4.9$^*$& 24.1$^*$& 4.2$^*$& n.a.& n.a.\\
AnnealedVAE \cite{burgess2018understanding}    &7.8$^*$& 10.7$^*$& 4.6$^*$& 6.7$^*$& n.a.& n.a.\\
FactorVAE \cite{kim2018disentangling}      &17.4& 14.3& \textbf{25.3}& 9.0& 34.7& 11.1\\
\midrule
D-VAE          &17.4& 9.4& 19.0& 8.5& 28.8& 12.8\\
FactorDVAE    &\textbf{21.7}& \textbf{15.5}& 23.2& \textbf{14.9}& \textbf{42.4}& \textbf{30.5}\\
\midrule
Selection      &39.5 &20.0 &22.7 &19.1 &40.1 &32.3\\
\bottomrule
\end{tabular}
\end{center}
\end{table*}
%
%
%
%%%%%%%%%%%%%%%%%%%%%%%%%%%%%%%%%%%%%%%%%%%%%%%%%%%%%%%%%%%%%%%%
%%%%%%%%%%%%%%%%%%%%%%%%%%%%%%%%%%%%%%%%%%%%%%%%%%%%%%%%%%%%%%%%
% Related work
%%%%%%%%%%%%%%%%%%%%%%%%%%%%%%%%%%%%%%%%%%%%%%%%%%%%%%%%%%%%%%%%
%%%%%%%%%%%%%%%%%%%%%%%%%%%%%%%%%%%%%%%%%%%%%%%%%%%%%%%%%%%%%%%%
%
\section{Related Work}
Previous studies have proposed various methods for utilizing discrete latent spaces.
The REINFORCE algorithm \cite{williams1992simple} utilizes the log derivative trick.
The Straight-Through estimator \cite{bengio2013estimating} back-propagates through hard samples by replacing the threshold function with the identity in the backward pass.
Additional prior work employed the nearest neighbor look-up called vector quantization \cite{van2017neural} to discretize the latent space.
Other approaches use reparameterization tricks \cite{kingma2013auto} that enable the gradient computation by removing the dependence of the density on the input parameters.
Maddison et al. \cite{maddison2017concrete} and Jang et al. \cite{jang2017categorical} propose the Gumbel-Softmax trick, a continuous %(but biased) 
reparameterization trick for categorical distributions.
Extensions of the Gumbel-Softmax trick discussed control variates \cite{tucker2017rebar,grathwohl2018backpropagation}, the local reparameterization trick \cite{shayer2018learning}, or the behavior of multiple sequential discrete components \cite{friede2021efficient}.
In this work, we focus on the structural impact of discrete representations on the latent space from the viewpoint of disentanglement.

\noindent State-of-the-art unsupervised disentanglement methods enhance Gaussian VAEs with various regularizers that encourage disentangling properties.
The $\beta$-VAE model \cite{higgins2017beta} introduces a hyperparameter to control the trade-off between the reconstruction loss and the KL-divergence term, promoting disentangled latent representations.
The annealedVAE \cite{burgess2018understanding} adapts to the $\beta$-VAE by annealing the $\beta$ hyperparameter during training.
FactorVAE \cite{kim2018disentangling} and $\beta$-TCVAE \cite{chen2018isolating} promote independence among latent variables by controlling the total correlation between them.
DIP-VAE-I and DIP-VAE-II \cite{kumar2017variational} are two variants that enforce disentangled latent factors by matching the covariance of the aggregated posterior to that of the prior.
Previous research has focused on augmenting the standard variational autoencoder with discrete factors \cite{makhzani2015adversarial,dupont2018learning,jeong2019learning} to improve disentangling properties.
In contrast, our goal is to replace the variational autoencoder with a categorical one, treating every ground-truth factor as a discrete representation.
%
%
%
%
%
%%%%%%%%%%%%%%%%%%%%%%%%%%%%%%%%
% Figure 4
%%%%%%%%%%%%%%%%%%%%%%%%%%%%%%%%
%
%
%
%
%%%%%%%%%%%%%%%%%%%%%%%%%%%%%%%%
% Figure 5
%%%%%%%%%%%%%%%%%%%%%%%%%%%%%%%%
%
\begin{figure}[t]
\begin{center}
\centerline{
    \includegraphics[width=0.49\columnwidth]{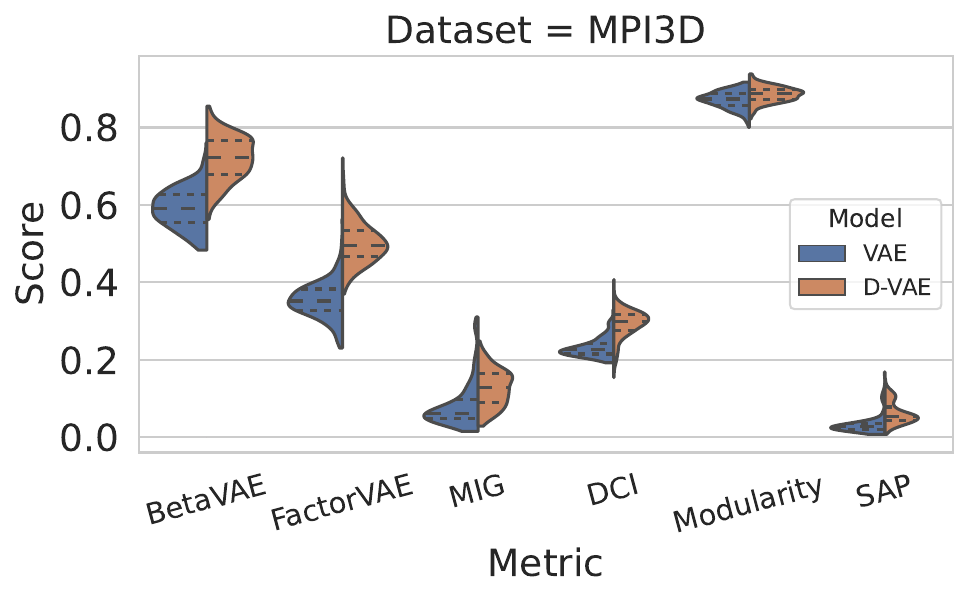}
    \includegraphics[width=0.49\columnwidth]{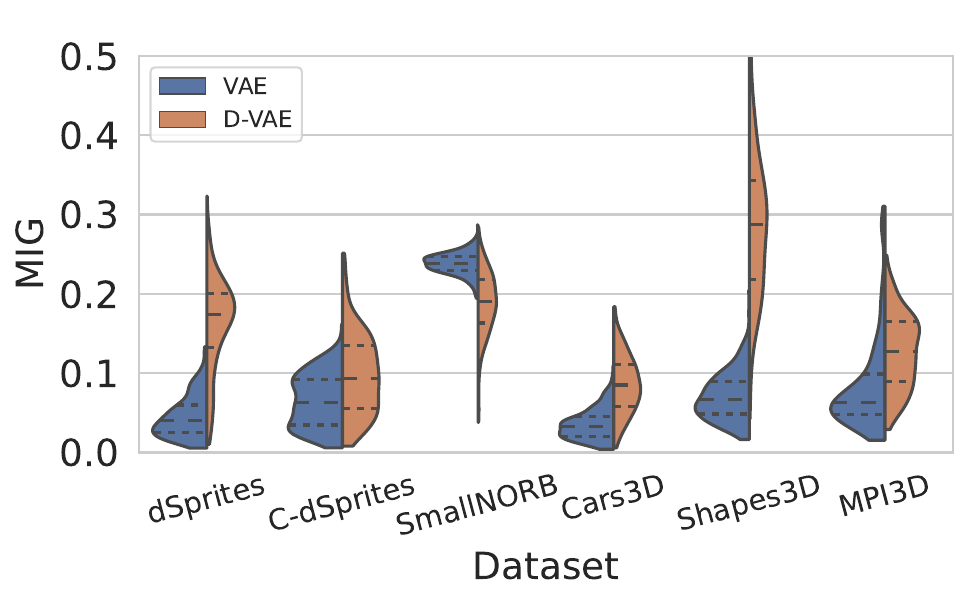}
}
\caption{Comparison between the unregularized Gaussian VAE and the discrete VAE by kernel density estimates of $300$ runs, respectively.
\textbf{Left:} Comparison on the MPI3D dataset w.r.t. the six disentanglement metrics.
The discrete model yields a better score for each metric, with median improvements ranging from $2$\% for Modularity to $104$\% for MIG.
\textbf{Right:} Comparison on all six datasets w.r.t. the MIG metric.
With the exception of SmallNORB, the discrete VAE yields a better score for all datasets with improvements of the median score ranging from $50$\% on C-dSprites to $336$\% on dSprites.
}%
\label{fig:unreg_mpi3d}%
\end{center}%
\end{figure}%
% %
% %
%
%
%%%%%%%%%%%%%%%%%%%%%%%%%%%%%%%%%%%%%%%%%%%%%%%%%%%%%%%%%%%%%%%%
%%%%%%%%%%%%%%%%%%%%%%%%%%%%%%%%%%%%%%%%%%%%%%%%%%%%%%%%%%%%%%%%
% Experimental setup
%%%%%%%%%%%%%%%%%%%%%%%%%%%%%%%%%%%%%%%%%%%%%%%%%%%%%%%%%%%%%%%%
%%%%%%%%%%%%%%%%%%%%%%%%%%%%%%%%%%%%%%%%%%%%%%%%%%%%%%%%%%%%%%%%
%
\section{Experimental Setup}
%
%
%
%%%%%%%%%%%%%%%%%%%%%%%%%%%%%%%%
% Methods
%%%%%%%%%%%%%%%%%%%%%%%%%%%%%%%%
%
\noindent\textbf{Methods.}
The experiments aim to compare the Gaussian VAE with the discrete VAE.
We consider the unregularized version and the total correlation penalizing method, VAE, D-VAE, FactorVAE \cite{kim2018disentangling} and FactorDVAE a version of FactorVAE for the D-VAE. We provide a detailed discussion of FactorDVAE in Appendix~\ref{app:improv}.
For the semi-supervised experiments, we augment each loss function with the supervised regularizer $R_s$ as in Appendix~\ref{app:improv}.
For the Gaussian VAE, we choose the BCE and the $L_2$ loss for $R_s$, respectively.
For the discrete VAE, we select the cross-entropy loss, once without and once with masked attention where we incorporate the knowledge about the number of unique variations.
We discuss the corresponding learning objectives in more detail in Appendix~\ref{app:improv}.

%
%
%
%%%%%%%%%%%%%%%%%%%%%%%%%%%%%%%%
% Datasets
%%%%%%%%%%%%%%%%%%%%%%%%%%%%%%%%
%
\noindent\textbf{Datasets.}
We consider six commonly used disentanglement datasets which offer explicit access to the ground-truth factors of variation: \emph{dSprites} \cite{higgins2017beta}, \emph{C-dSprites} \cite{locatello2019challenging}, \emph{SmallNORB} \cite{lecun2004learning}, \emph{Cars3D} \cite{reed2015deep}, \emph{Shapes3D} \cite{kim2018disentangling} and \emph{MPI3D} \cite{gondal2019transfer}.
We provide a more detailed description of the datasets in Table~\ref{tb:data} in Appendix~\ref{app:data}.
%
%
%
%%%%%%%%%%%%%%%%%%%%%%%%%%%%%%%%
% Metrics
%%%%%%%%%%%%%%%%%%%%%%%%%%%%%%%%
%

\noindent\textbf{Metrics.}
We consider the commonly used disentanglement metrics that have been discussed in detail in \cite{locatello2019challenging} to evaluate the representations: \emph{BetaVAE} metric \cite{higgins2017beta}, \emph{FactorVAE} metric \cite{kim2018disentangling}, \emph{Mutual Information Gap} (MIG) \cite{chen2018isolating}, \emph{DCI Disentanglement} (DCI) \cite{eastwood2018framework}, \emph{Modularity} \cite{ridgeway2018learning} and \emph{SAP score} (SAP) \cite{kumar2017variational}.
As illustrated on the right side of Figure~\ref{fig:st-gap_downstream}, the MIG score seems to be the most reliable indicator of sample efficiency across different datasets.
Therefore, we primarily focus on the MIG disentanglement score.
We discuss this in more detail in Appendix~\ref{app:further}.
%
%
%
%%%%%%%%%%%%%%%%%%%%%%%%%%%%%%%%
% Experimental protocol
%%%%%%%%%%%%%%%%%%%%%%%%%%%%%%%%
%

\noindent\textbf{Experimental protocol.}
We adopt the experimental setup of prior work (\cite{locatello2019challenging} and \cite{locatello2019disentangling}) for the unsupervised and for the semi-supervised experiments, respectively.
Specifically, we utilize the same neural architecture for all methods so that all differences solely emerge from the distribution of the type of VAE.
For the unsupervised case, we run each considered method on each dataset for $50$ different random seeds.
Since the two unregularized methods do not have any extra hyperparameters, we run them for $300$ different random seeds instead.
For the semi-supervised case, we consider two numbers ($100$/$1000$) of perfectly labeled examples and split the labeled examples ($90$\%/$10$\%) into a training and validation set.
We choose $6$ values for the correlation penalizing hyperparameter $\gamma$ and for the semi-supervising hyperparameter $\omega$ from Equation~\ref{eq:factor_dvae} and~\ref{eq:semi-sup} in Appendix~\ref{app:improv}, respectively.
We present the full implementation details in Appendix~\ref{app:impl}.
%
%
%
%
%
%
%
%%%%%%%%%%%%%%%%%%%%%%%%%%%%%%%%
% Figure 6
%%%%%%%%%%%%%%%%%%%%%%%%%%%%%%%%
%
\begin{figure}[t]
\begin{center}
\centerline{
    \includegraphics[width=0.51\columnwidth]{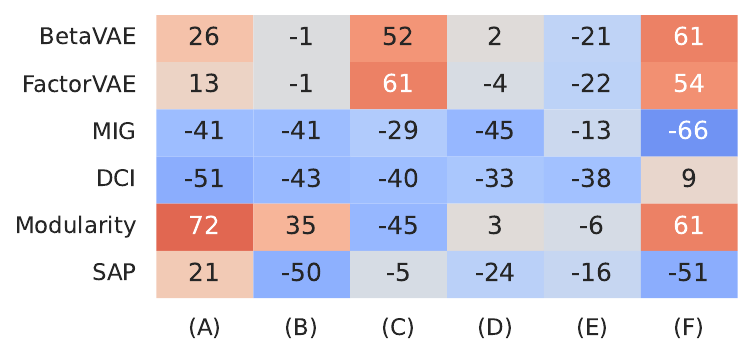}
    \includegraphics[width=0.48\columnwidth]{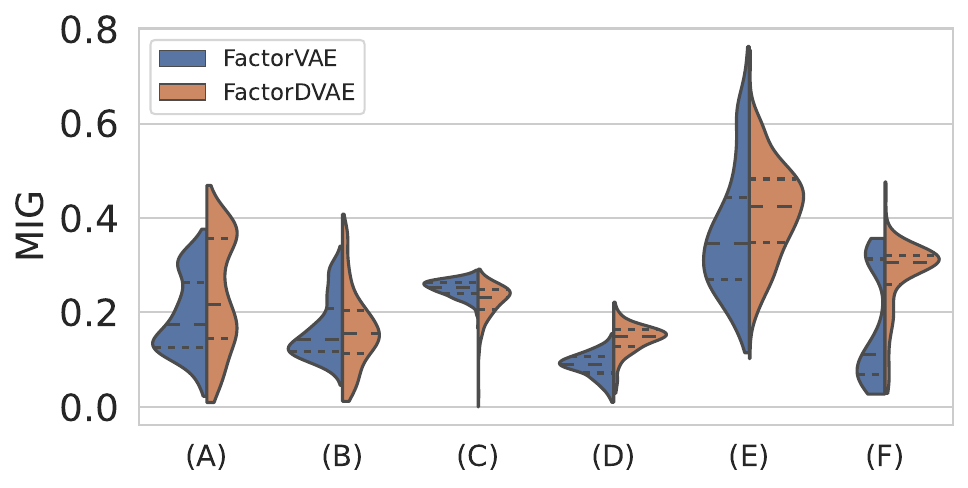}
}
\caption{Disentangling properties of FactorDVAE on different datasets: dSprites (A), C-dSprites (B), SmallNORB (C), Cars3D (D), Shapes3D (E), MPI3D (F).
\textbf{Left:} The Spearman rank correlation between various disentangling metrics and $\operatorname{Gap}_{ST}$ of D-VAE and FactorDVAE combined.
A small $\operatorname{Gap}_{ST}$ indicates high disentangling scores for most datasets regarding the MIG, DCI, and SAP metrics.
\textbf{Right:} A comparison of the total correlation regularizing Gaussian and the discrete model w.r.t. the MIG metric.
The discrete model yields a better score for all datasets but SmallNORB with median improvements ranging from $8$\% on C-dSprites to $175$\% on MPI3D.}
\label{fig:st-gap_factor}
\end{center}
\end{figure}
%
%
%
%
%%%%%%%%%%%%%%%%%%%%%%%%%%%%%%%%%%%%%%%%%%%%%%%%%%%%%%%%%%%%%%%%
%%%%%%%%%%%%%%%%%%%%%%%%%%%%%%%%%%%%%%%%%%%%%%%%%%%%%%%%%%%%%%%%
% Experiments
%%%%%%%%%%%%%%%%%%%%%%%%%%%%%%%%%%%%%%%%%%%%%%%%%%%%%%%%%%%%%%%%
%%%%%%%%%%%%%%%%%%%%%%%%%%%%%%%%%%%%%%%%%%%%%%%%%%%%%%%%%%%%%%%%
%
\section{Experimental Results}
First, we investigate whether a discrete VAE offers advantages over Gaussian VAEs in terms of disentanglement properties, finding that the discrete model generally outperforms its Gaussian counterpart and showing that the FactorDVAE achieves new state-of-the-art MIG scores on most datasets.
Additionally, we propose a model selection criterion based on $\operatorname{Gap}_{ST}$ to find good discrete models solely using unsupervised scores.
Lastly, we examine how incorporating label information can further enhance discrete representations.
The implementations are in JAX and Haiku and were run on a RTX A6000 GPU.\footnote{The implementations and Appendix are at \url{https://github.com/david-friede/lddr}.}
%
%
%
%%%%%%%%%%%%%%%%%%%%%%%%%%%%%%%%
% Backbone Comparison
%%%%%%%%%%%%%%%%%%%%%%%%%%%%%%%%
%
\subsection{Improvement in unsupervised disentanglement properties}\label{ex:backbone}
\noindent\textbf{Comparison of the unregularized models.}
In the first experiment, we aim to answer our main research question of whether discrete latent spaces yield structural advantages over their Gaussian counterparts.
Figure~\ref{fig:unreg_mpi3d} depicts the comparison regarding the disentanglement scores (left) and the datasets (right).
The discrete model achieves a better score on the MPI3D dataset for each metric with median improvements ranging from $2$\% for Modularity to $104$\% for MIG.
Furthermore, the discrete model yields a better score for all datasets but SmallNORB with median improvements ranging from $50$\% on C-dSprites to $336$\% on dSprites.
More detailed results can be found in Table~\ref{tb:quant}, Figure~\ref{fig:app_unreg_all_1}, and Figure~\ref{fig:app_unreg_all_2} in Appendix~\ref{app:exp}.
Taking into account all datasets and metrics, the discrete VAE improves over its Gaussian counterpart in $31$ out of $36$ cases.

\noindent\textbf{Comparison of the total correlation regularizing models.}
For each VAE, we choose the same $6$ values of hyperparameter $\gamma$ for the total correlation penalizing method and train $50$ copies, respectively.
The right side of Figure~\ref{fig:st-gap_factor} depicts the comparison of FactorVAE and FactorDVAE w.r.t. the MIG metric.
The discrete model achieves a better score for all datasets but SmallNORB with median improvements ranging from $8$\% on C-dSprites to $175$\% on MPI3D.
%
%
%
%%%%%%%%%%%%%%%%%%%%%%%%%%%%%%%%
% Figure 7
%%%%%%%%%%%%%%%%%%%%%%%%%%%%%%%%
%
\begin{figure}[t]
\begin{center}
\centerline{
    \includegraphics[width=0.245\columnwidth]{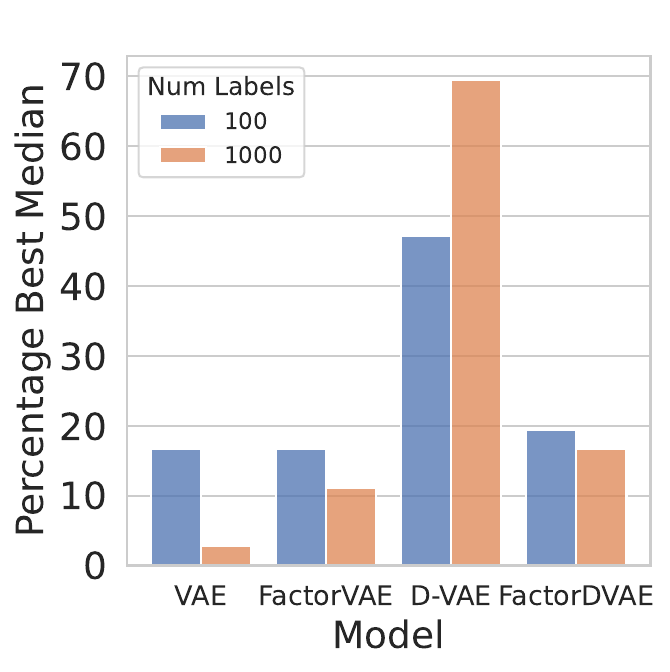}
    \includegraphics[width=0.245\columnwidth]{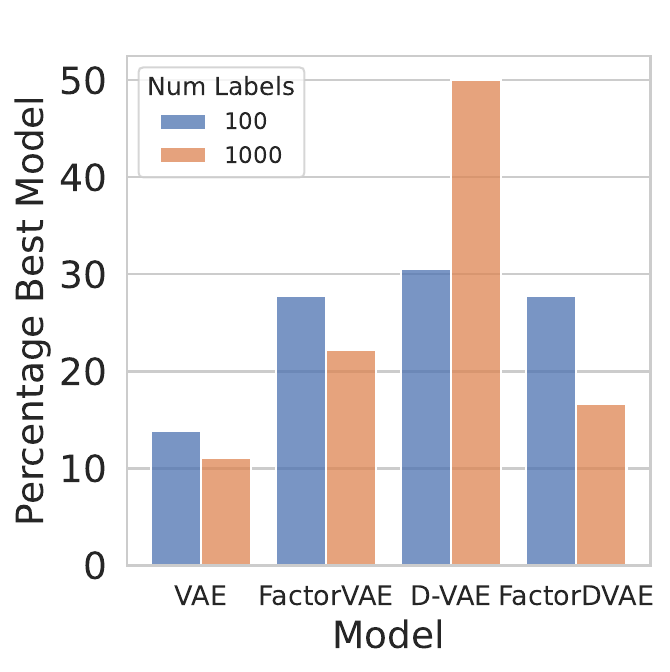}
    \includegraphics[width=0.245\columnwidth]{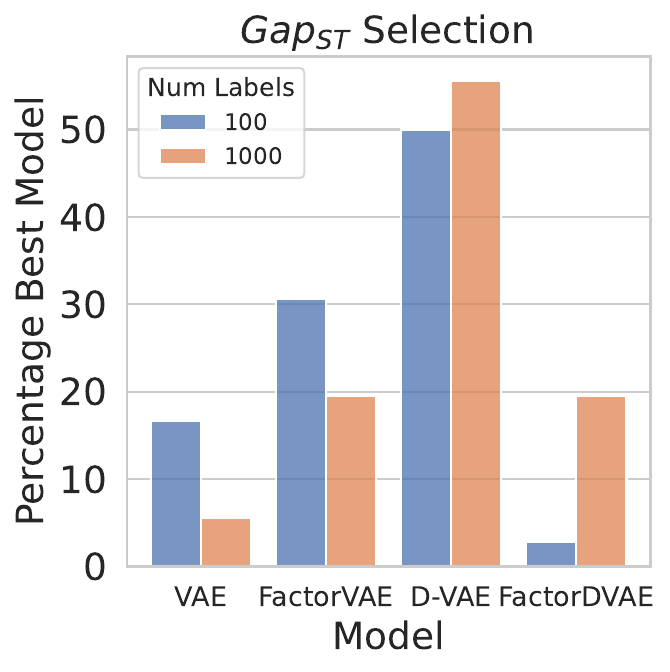}
    \includegraphics[width=0.245\columnwidth]{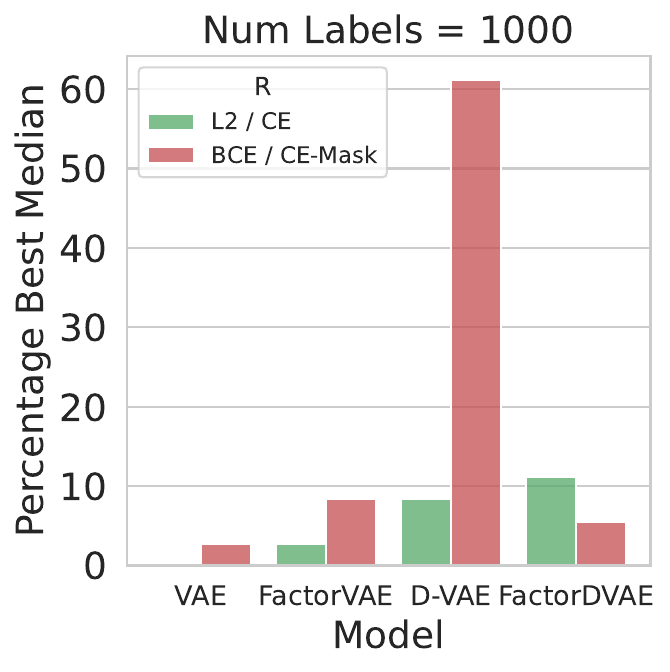}
}
\caption{The percentage of each semi-supervised method being the best over all datasets and disentanglement metrics for different selection methods: median, lowest $R_s$, lowest  $\operatorname{Gap}_{ST}$, median for $1000$ labels.
The unregularized discrete method outperforms the other methods in semi-supervised disentanglement task.
Utilizing the masked regularizer improves over the unmasked one.}
\label{fig:semisupervised}
\end{center}
\end{figure}
%
%
%
%
%%%%%%%%%%%%%%%%%%%%%%%%%%%%%%%%
% Model Improvements
%%%%%%%%%%%%%%%%%%%%%%%%%%%%%%%%
%
\subsection{Match state-of-the-art unsupervised disentanglement methods}
Current state-of-the-art unsupervised disentanglement methods enrich Gaussian VAEs with various regularizers encouraging disentangling properties.
Table~\ref{tb:mig} depicts the MIG scores of all methods as reported in \cite{locatello2019challenging} utilizing the same architecture as us.
FactorDVAE achieves new state-of-the-art MIG scores on all datasets but SmallNORB, improving the previous best scores by over $17$\% on average.
These findings suggest that incorporating results from the disentanglement literature might lead to even stronger models based on discrete representations.
%
%
%
%%%%%%%%%%%%%%%%%%%%%%%%%%%%%%%%
% Unsupervised Model Selection
%%%%%%%%%%%%%%%%%%%%%%%%%%%%%%%%
%
\subsection{Unsupervised selection of models with strong disentanglement} 
A remaining challenge in the disentanglement literature is selecting the hyperparameters and random seeds that lead to good disentanglement scores \cite{locatello2019disentangling}.
We propose a model selection based on an unsupervised score measuring the discreteness of the latent space utilizing $\operatorname{Gap}_{ST}$ from Section~\ref{sec:gap_st}.
The left side of Figure~\ref{fig:st-gap_factor} depicts the Spearman rank correlation between various disentangling metrics and $\operatorname{Gap}_{ST}$ of D-VAE and FactorDVAE combined.
Note that the unregularized D-VAE model can be identified as a FactorDVAE model with $\gamma=0$.
A small Straight-Through Gap corresponds to high disentangling scores for most datasets regarding the MIG, DCI, and SAP metrics.
This correlation is most vital for the MIG metric.
We anticipate finding good hyperparameters by selecting those models yielding the smallest $\operatorname{Gap}_{ST}$.
The last row of Table~\ref{tb:mig} confirms this finding.
This model selection yields MIG scores that are, on average, $22$\% better than the median score and not worse than $6$\%.
%
%
%
%%%%%%%%%%%%%%%%%%%%%%%%%%%%%%%%
% Unsupervised Model Selection
%%%%%%%%%%%%%%%%%%%%%%%%%%%%%%%%
%
\subsection{Utilize label information to improve discrete representations}
Locatello et al. \cite{locatello2019disentangling} employ the semi-supervised regularizer $R_s$ by including $90$\% of the label information during training and utilizing the remaining $10$\% for a model selection.
We also experiment with a model selection based on the $\operatorname{Gap}_{ST}$ value.
Figure~\ref{fig:semisupervised} depicts the percentage of each semi-supervised method being the best over all datasets and disentanglement metrics.
The unregularized discrete method surpasses the other methods on the semi-supervised disentanglement task.
The advantage of the discrete models is more significant for the median values than for the model selection.
Utilizing $\operatorname{Gap}_{ST}$ for selecting the discrete models only partially mitigates this problem.
Incorporating the number of unique variations by utilizing the masked regularizer improves the disentangling properties significantly, showcasing another advantage of the discrete latent space.
The quantiles of the discrete models can be found in Table~\ref{tb:quant_semi} in Appendix~\ref{app:exp}.
\subsection{Visualization of the latent categories}
\begin{figure}[t]
\begin{center}
\centerline{
    \begin{tikzpicture}[>=latex, node distance = 1.18cm, thick]
        \newcommand{\widthFactor}{.095}
        \node[inner sep=0pt, label={[label distance=-0.05cm]90:\textbf{\scriptsize Input}}] (1) at (0,0)
        {\includegraphics[width=\widthFactor\linewidth]{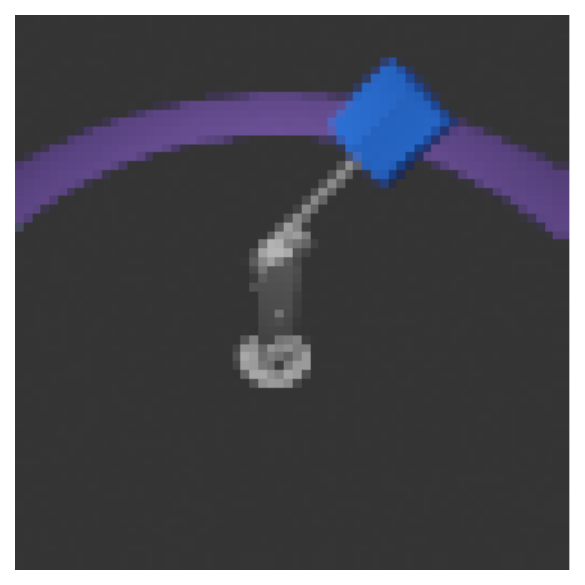}};
        \node[inner sep=0pt, below of=1] (2) 
        {\includegraphics[width=\widthFactor\linewidth]{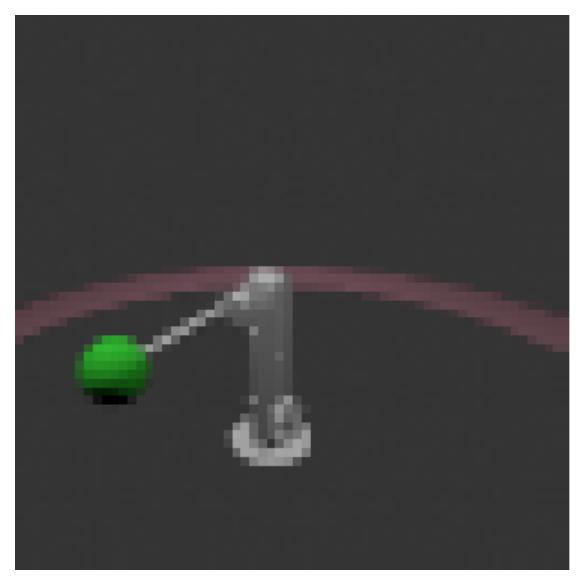}};
        \node[inner sep=0pt, below of=2] (3)
        {\includegraphics[width=\widthFactor\linewidth]{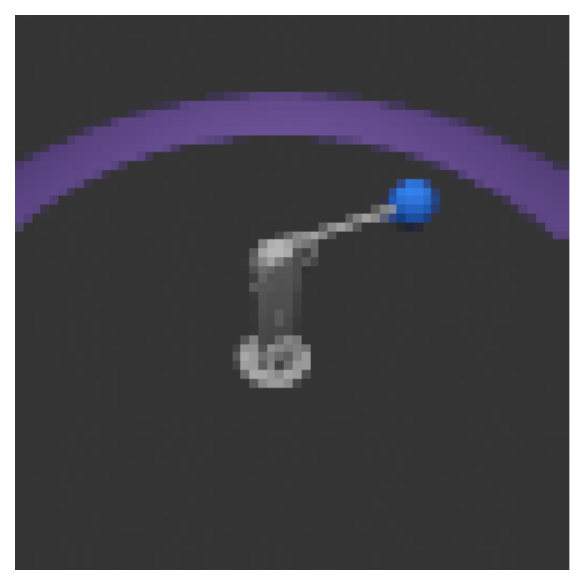}};
        \node[inner sep=0pt, below of=3] (4)
        {\includegraphics[width=\widthFactor\linewidth]{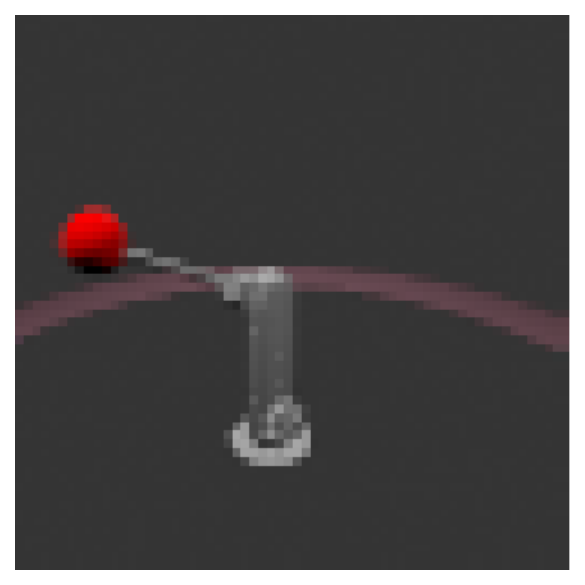}};
        \node[inner sep=0pt, right of=1, label=90:\textbf{\scriptsize Recons}] (1r)
        {\includegraphics[width=\widthFactor\linewidth]{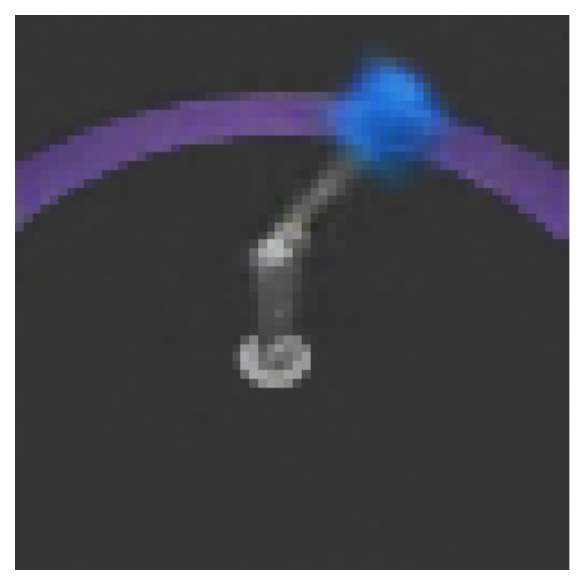}};
        \node[inner sep=0pt, below of=1r] (2r) 
        {\includegraphics[width=\widthFactor\linewidth]{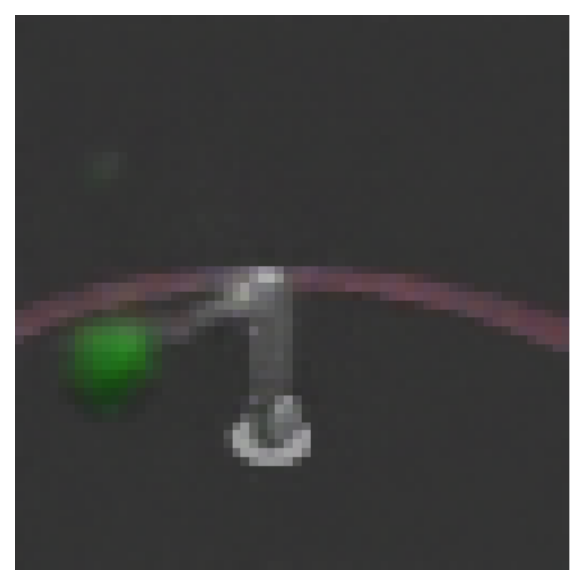}};
        \node[inner sep=0pt, below of=2r] (3r)
        {\includegraphics[width=\widthFactor\linewidth]{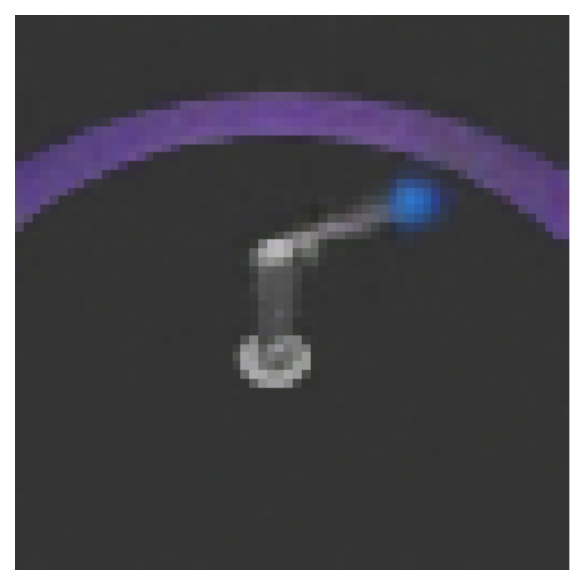}};
        \node[inner sep=0pt, below of=3r] (4r)
        {\includegraphics[width=\widthFactor\linewidth]{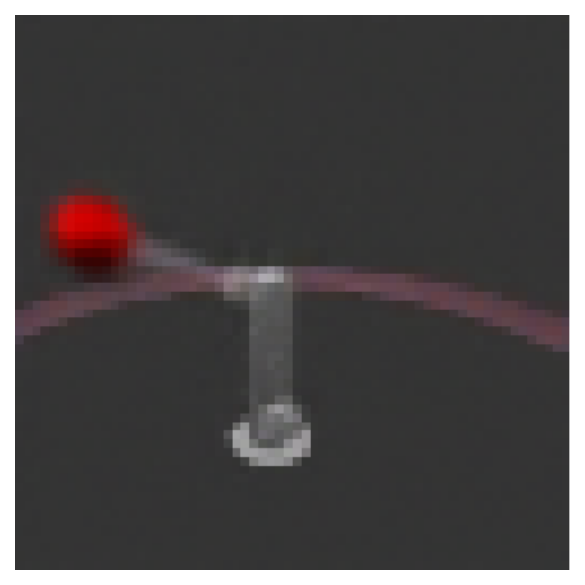}};
        \node[inner sep=0pt, right = 0.5cm of 1r] (1a)
        {\includegraphics[width=\widthFactor\linewidth]{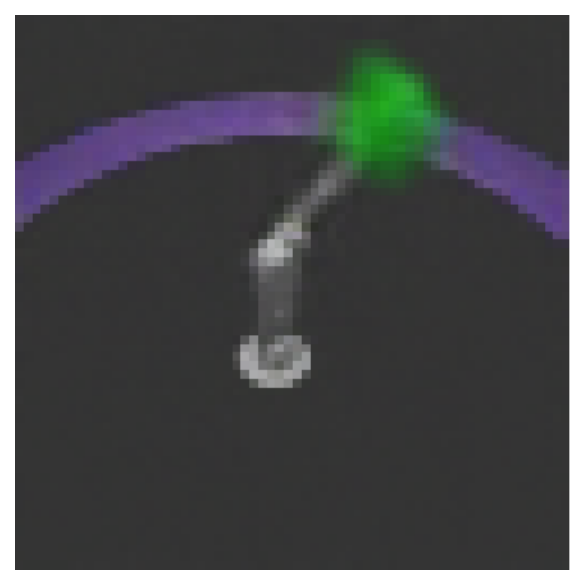}};
        \node[inner sep=0pt, below of=1a] (2a) 
        {\includegraphics[width=\widthFactor\linewidth]{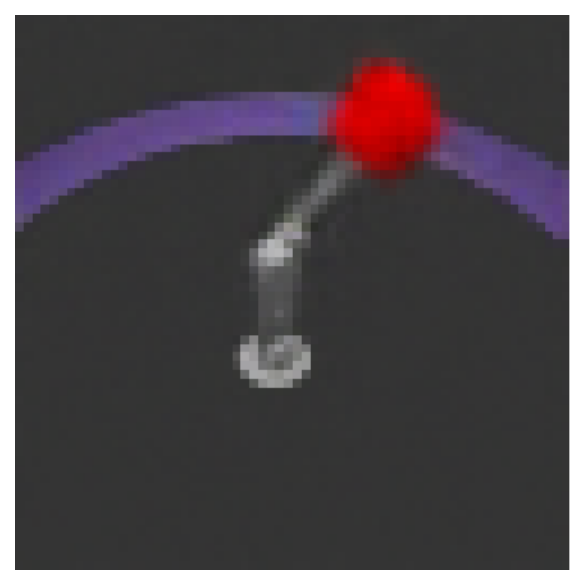}};
        \node[inner sep=0pt, below of=2a] (3a)
        {\includegraphics[width=\widthFactor\linewidth]{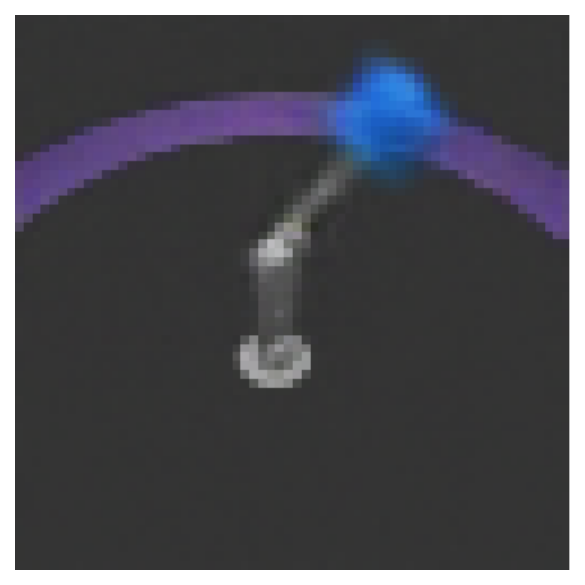}};
        \node[inner sep=0pt, below of=3a, label={[label distance=-1.5cm]90:{\scriptsize Color}}] (4a)
        {\includegraphics[width=\widthFactor\linewidth]{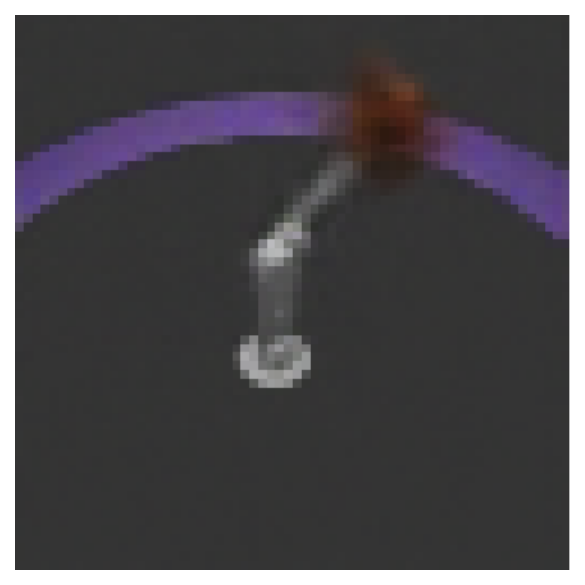}};
        \node[inner sep=0pt, right of = 1a] (1b)
        {\includegraphics[width=\widthFactor\linewidth]{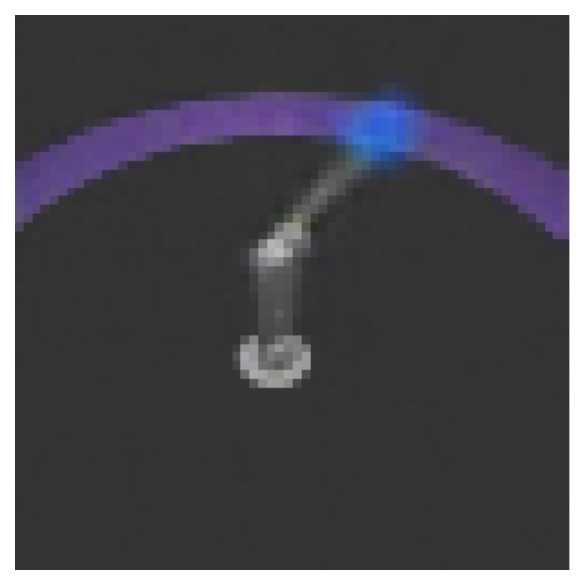}};
        \node[inner sep=0pt, below of=1b] (2b) 
        {\includegraphics[width=\widthFactor\linewidth]{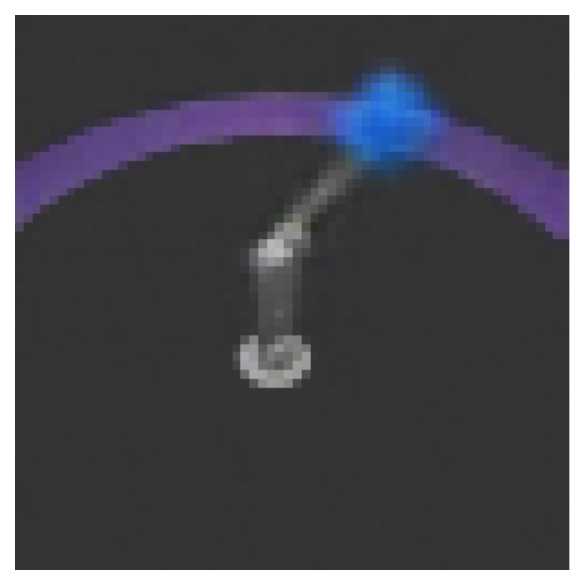}};
        \node[inner sep=0pt, below of=2b] (3b)
        {\includegraphics[width=\widthFactor\linewidth]{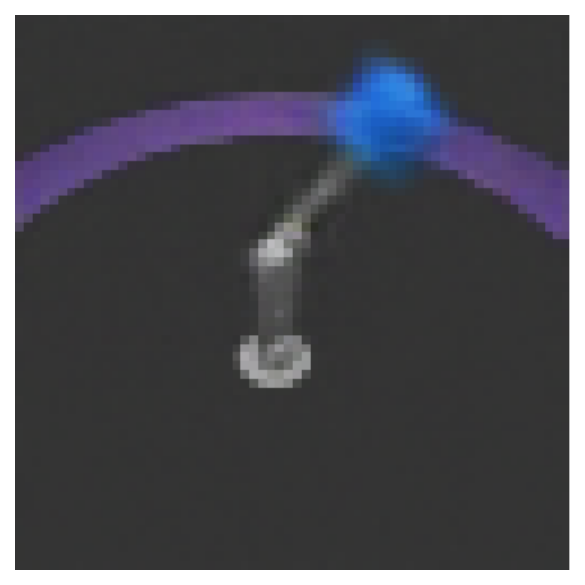}};
        \node[inner sep=0pt, below of=3b, label={[label distance=-1.55cm]90:{\scriptsize Shape}}] (4b)
        {\includegraphics[width=\widthFactor\linewidth]{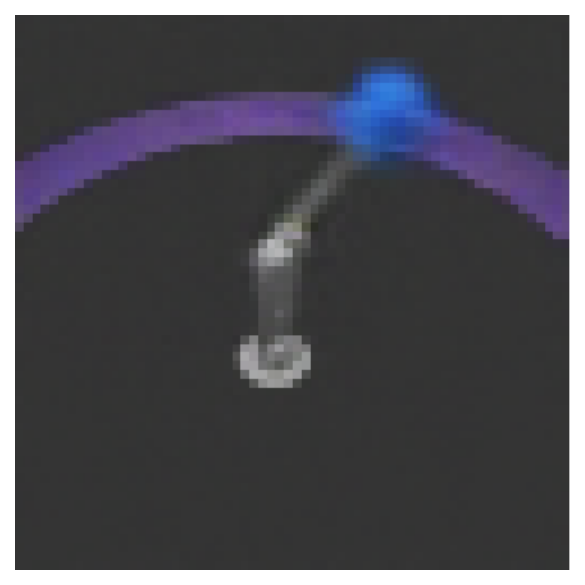}};
        \node[inner sep=0pt, right of = 1b] (1c)
        {\includegraphics[width=\widthFactor\linewidth]{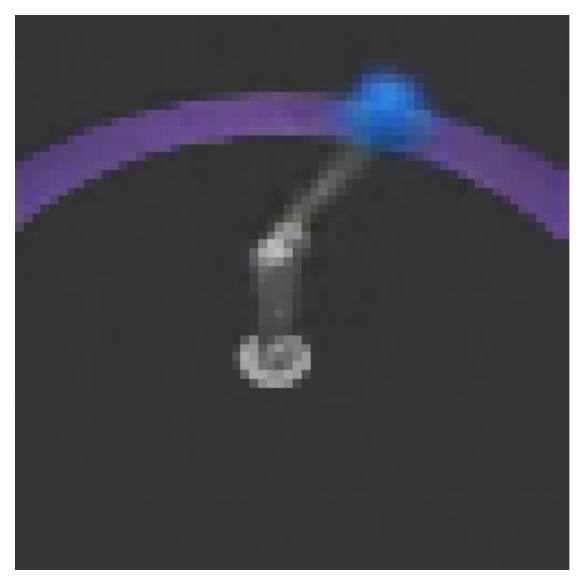}};
        \node[inner sep=0pt, below of=1c, label={[label distance=-1.5cm]90:{\scriptsize Size}}] (2c)
        {\includegraphics[width=\widthFactor\linewidth]{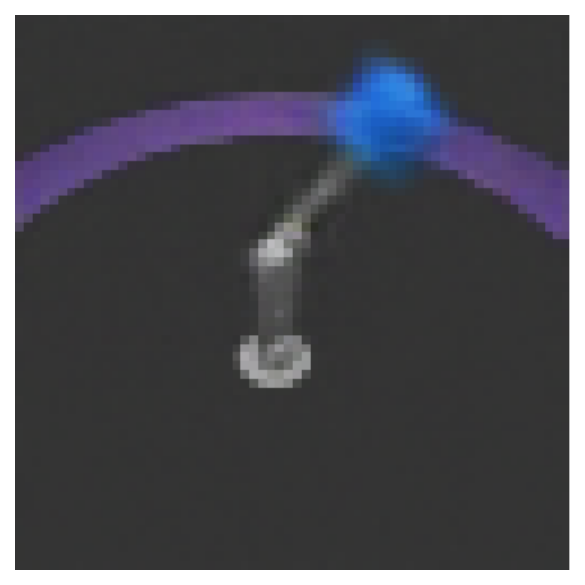}};
        \node[inner sep=0pt, right of = 1c, label={[label distance=-0.06cm]90:\textbf{\scriptsize Latent Category Traversals}}] (1d)
        {\includegraphics[width=\widthFactor\linewidth]{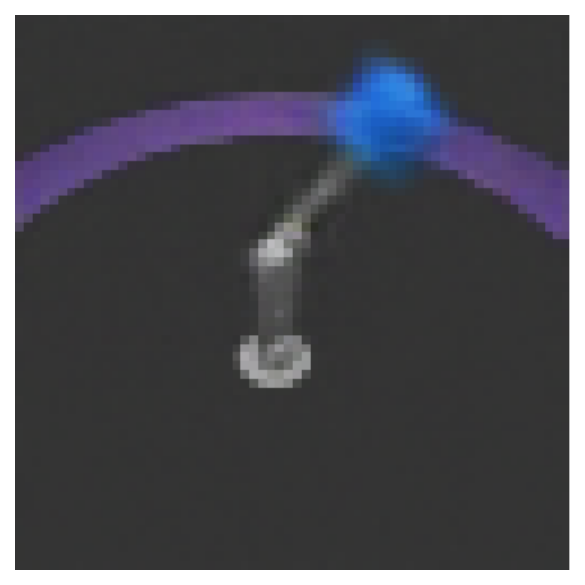}};
        \node[inner sep=0pt, below of = 1d] (2d)
        {\includegraphics[width=\widthFactor\linewidth]{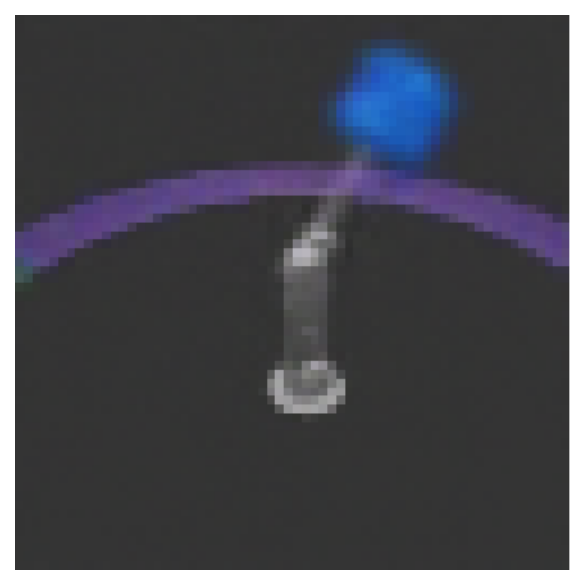}};        
        \node[inner sep=0pt, below of = 2d, label={[label distance=-1.53cm]90:{\scriptsize Angle}}] (3d)
        {\includegraphics[width=\widthFactor\linewidth]{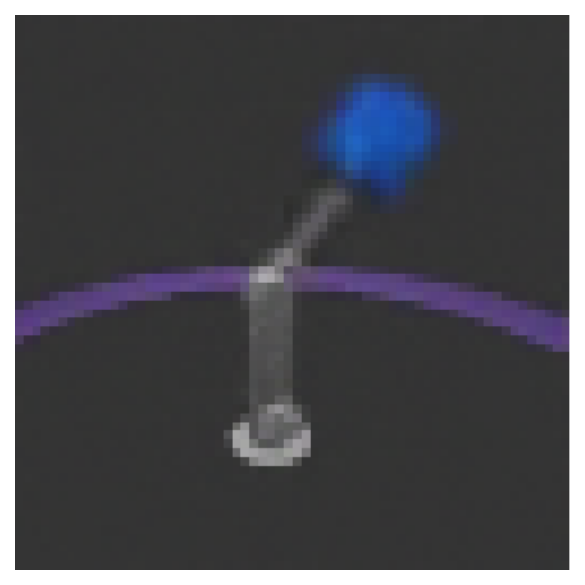}}; 
        \node[inner sep=0pt, right of = 1d] (1e)
        {\includegraphics[width=\widthFactor\linewidth]{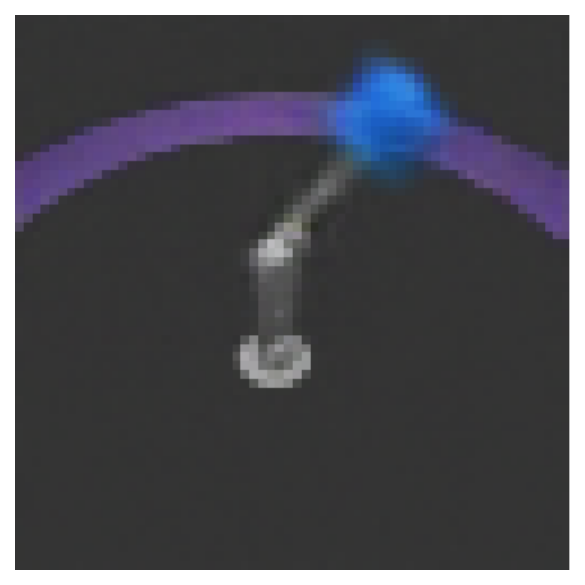}};
        \node[inner sep=0pt, below of = 1e] (2e)
        {\includegraphics[width=\widthFactor\linewidth]{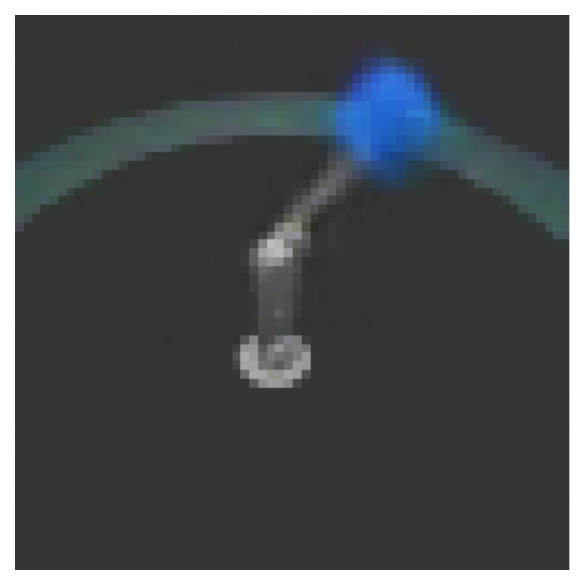}};        
        \node[inner sep=0pt, below of = 2e, label={[label distance=-1.5cm]90:{\scriptsize BG}}] (3e)
        {\includegraphics[width=\widthFactor\linewidth]{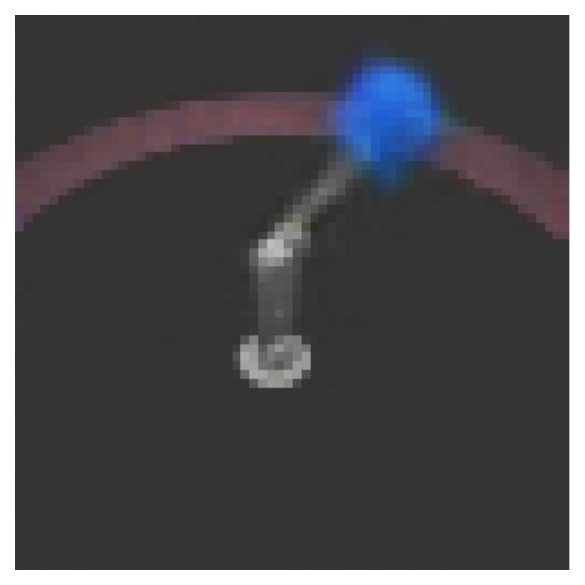}};
        \node[inner sep=0pt, right of = 1e] (1f)
        {\includegraphics[width=\widthFactor\linewidth]{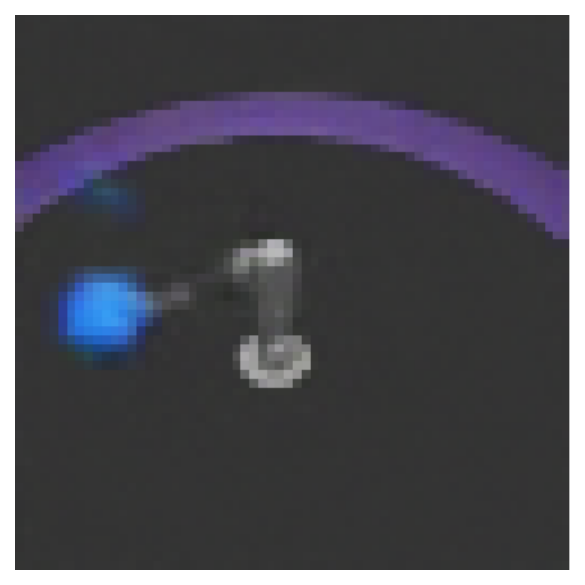}};
        \node[inner sep=0pt, below of=1f] (2f) 
        {\includegraphics[width=\widthFactor\linewidth]{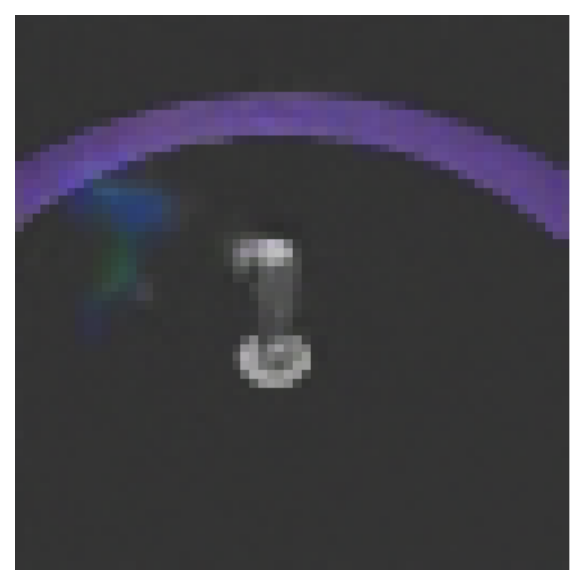}};
        \node[inner sep=0pt, below of=2f] (3f)
        {\includegraphics[width=\widthFactor\linewidth]{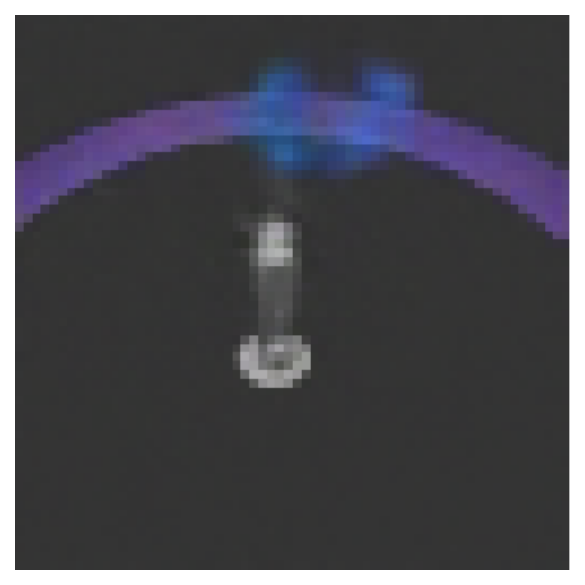}};
        \node[inner sep=0pt, below of=3f, label={[label distance=-1.5cm]90:{\scriptsize DOF1}}] (4f)
        {\includegraphics[width=\widthFactor\linewidth]{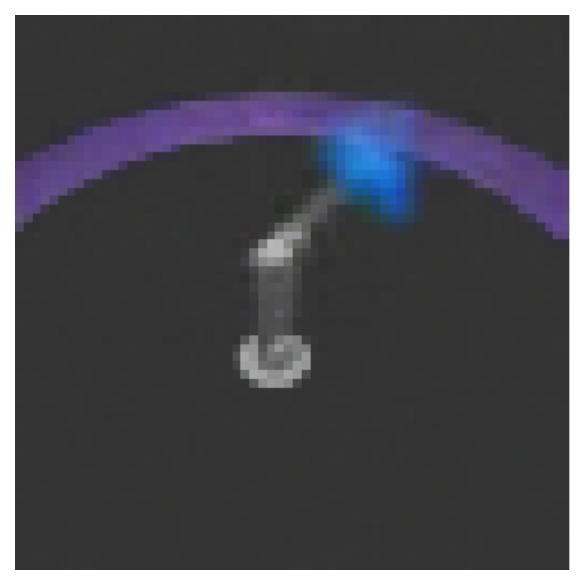}};    
        \node[inner sep=0pt, right of = 1f] (1g)
        {\includegraphics[width=\widthFactor\linewidth]{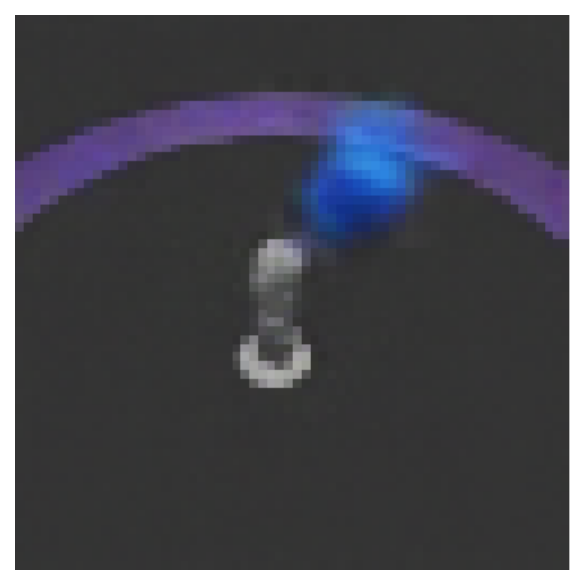}};
        \node[inner sep=0pt, below of=1g] (2g) 
        {\includegraphics[width=\widthFactor\linewidth]{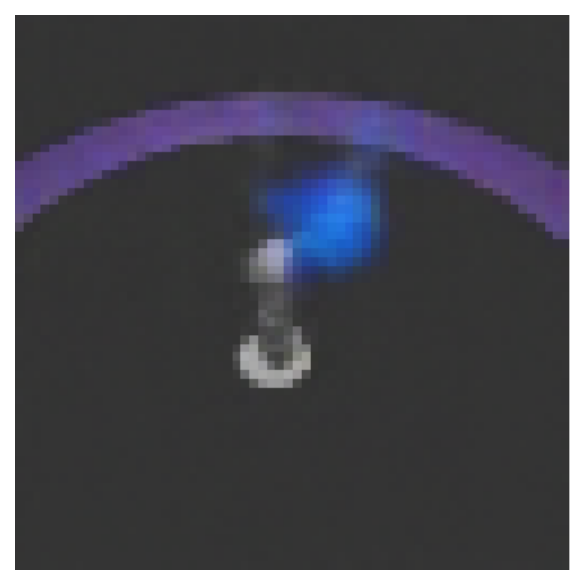}};
        \node[inner sep=0pt, below of=2g] (3g)
        {\includegraphics[width=\widthFactor\linewidth]{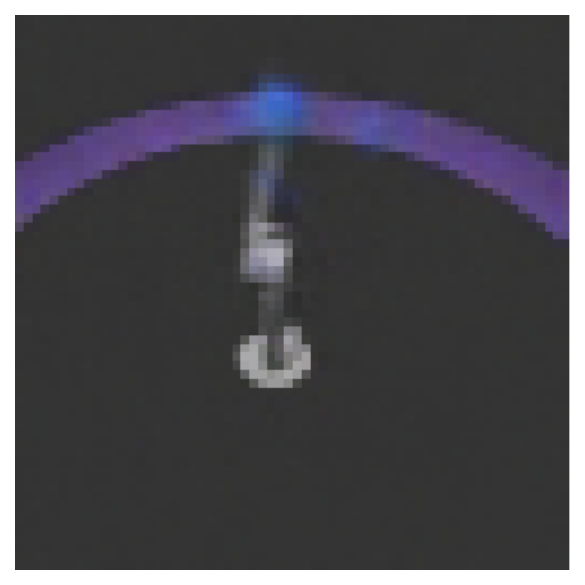}};
        \node[inner sep=0pt, below of=3g, label={[label distance=-1.5cm]90:{\scriptsize DOF2}}] (4g)
        {\includegraphics[width=\widthFactor\linewidth]{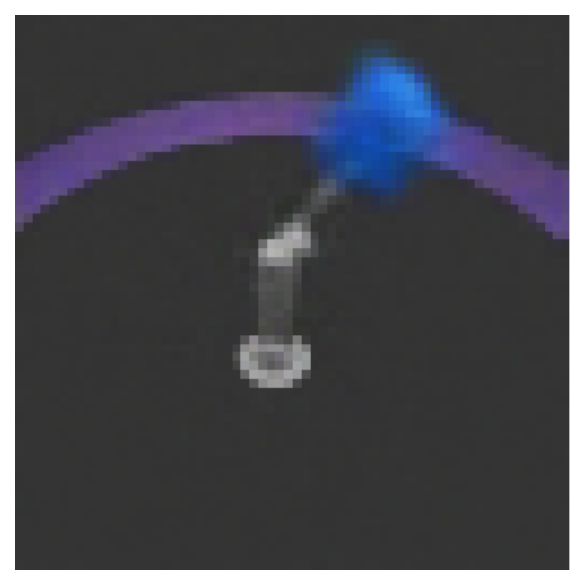}};
        \begin{scope}[transform canvas={xshift=0.85cm}]
            \draw[->] (1g.center) -- (4g.center);
        \end{scope}
    \end{tikzpicture}
}
\caption{%
Reconstructions and latent space traversals of the semi-supervised D-VAE, utilizing masked attentions with the lowest $R_s$ value. The masked attention allows for the incorporation of the number of unique variations, such as two for the object size. We visualize four degrees of freedom (DOF), selected equidistantly from the total of $40$.
\textbf{Left:}~The reconstructions are easily recognizable, albeit with blurry details.
\textbf{Right:}~The object color, size, camera angle, and background color (BG) are visually disentangled.
The object shape and the DOF factors remain partially entangled.}
\label{fig:recon}%
\end{center}
\end{figure}

Prior work uses latent space traversals for qualitative analysis of representations \cite{higgins2017beta,burgess2018understanding,kim2018disentangling,watters2019spatial}. 
A latent vector $\bm{z} \sim q_{\phi}(\bm{z}|\bm{x})$ is sampled, and each dimension $z_i$ is traversed while keeping the other dimensions constant.
The traversals are then reconstructed and visualized.
Unlike the Gaussian case, the D-VAE's latent space is known beforehand, allowing straightforward traversal along the categories.
Knowing the number of unique variations lets us use masked attention to determine the number of each factor's categories, improving latent space interpretability.
Figure~\ref{fig:recon} illustrates the reconstructions of four random inputs and latent space traversals of the semi-supervised D-VAE utilizing masked attentions.
While the reconstructions are easily recognizable, their details can be partially blurry, particularly concerning the object shape.
The object color, object size, camera angle, and background color are visually disentangled, and their categories can be selected straightforwardly to create targeted observations.
%
%
%
%%%%%%%%%%%%%%%%%%%%%%%%%%%%%%%%
% Figure 
%%%%%%%%%%%%%%%%%%%%%%%%%%%%%%%%
%
%
%
%
%
%
%%%%%%%%%%%%%%%%%%%%%%%%%%%%%%%%%%%%%%%%%%%%%%%%%%%%%%%%%%%%%%%%
%%%%%%%%%%%%%%%%%%%%%%%%%%%%%%%%%%%%%%%%%%%%%%%%%%%%%%%%%%%%%%%%
% Conclusion
%%%%%%%%%%%%%%%%%%%%%%%%%%%%%%%%%%%%%%%%%%%%%%%%%%%%%%%%%%%%%%%%
%%%%%%%%%%%%%%%%%%%%%%%%%%%%%%%%%%%%%%%%%%%%%%%%%%%%%%%%%%%%%%%%
%
\section{Conclusion}
In this study, we investigated the benefits of discrete latent spaces in the context of learning disentangled representations by examining the effects of substituting the standard Gaussian VAE with a categorical VAE. Our findings revealed that the underlying grid structure of categorical distributions mitigates the rotational invariance issue associated with multivariate Gaussian distributions, thus serving as an efficient inductive prior for disentangled representations.

In multiple experiments, we demonstrated that categorical VAEs outperform their Gaussian counterparts in disentanglement.
We also determined that the categorical VAE provides an unsupervised score, the Straight-Through Gap, which correlates with some disentanglement metrics, providing, to the best of our knowledge, the first unsupervised  model selection score for disentanglement.

However, our study has limitations.
We focused on discrete latent spaces, without investigating the impact of vector quantization on disentanglement.
Furthermore, the Straight-Through Gap does not show strong correlation with disentanglement scores, affecting model selection accuracy.
Additionally, our reconstructions can be somewhat blurry and may lack quality.

Our results offer a promising direction for future research in developing more powerful models with discrete latent spaces. Such future research could incorporate findings from the disentanglement literature and potentially develop novel regularizations tailored to discrete latent spaces.

%% file: appendix.tex
%
%
%
%%%%%%%%%%%%%%%%%%%%%%%%%%%%%%%%%%%%%%%%%%%%%%%%%%%%%%%%%%%%%%%%
%%%%%%%%%%%%%%%%%%%%%%%%%%%%%%%%%%%%%%%%%%%%%%%%%%%%%%%%%%%%%%%%
% Appendix
%%%%%%%%%%%%%%%%%%%%%%%%%%%%%%%%%%%%%%%%%%%%%%%%%%%%%%%%%%%%%%%%
%%%%%%%%%%%%%%%%%%%%%%%%%%%%%%%%%%%%%%%%%%%%%%%%%%%%%%%%%%%%%%%%
%
%
%
%
\newpage
\begin{subappendices}
\renewcommand{\thesection}{\arabic{section}}%
%
%
%
%%%%%%%%%%%%%%%%%%%%%%%%%%%%%%%%
% Proofs
%%%%%%%%%%%%%%%%%%%%%%%%%%%%%%%%
%
\section{Proofs} \label{app:proof}
\subsubsection{Proof of Proposition~\ref{prop:fn_f}}
\begin{proof}
  For the sake of clarity, we ignore the $i$-index in our notation and write the $j$-index as a subscript.\\
  Part (a):
  Let $J$ be the set of all indices of $\bm{\alpha}$ with $\alpha_j=0$ and let $m^\prime=m-|J|$ be the number of elements of $\bm{\alpha}$ that are non-zero.
  We will first show that
  \begin{equation*}
    \operatorname{supp}\bigl(\operatorname{GS}(\bm{\alpha})\bigr)=\operatorname{int}\{\bm{y}\in\mathbb{R}^n\;|\;y_j\in[0,1],\sum_{j=1}^m y_j = 1, y_k = 0 \text{ for } k\in J\}.
  \end{equation*}
  Let $P_{\bm{\alpha}}:\mathbb{R}^m\rightarrow\mathbb{R}^{m^\prime}$ be the projection that maps $\bm{\alpha}$ on its non-zero elements $\bm{\alpha}^\prime = P_{\bm{\alpha}}(\bm{\alpha})$ with $\alpha^{\prime}_j\neq0$ for all $j\in[m^\prime]$.
  We write $ P_{\bm{\alpha}}^{-1}(\bm{\alpha}^\prime) = \bm{\alpha}$ for the inverse of the projection.
  Sampling  $z \sim \operatorname{GS}(\bm{\alpha})$ is then defined by $P_{\bm{\alpha}}^{-1}(\bm{z}^\prime)$ for $\bm{z}^\prime \sim \operatorname{GS}(\bm{\alpha}^\prime)$.
  By Maddison et. al \cite{maddison2017concrete}, Proposition 1a, we know that the density of $\operatorname{GS}(\bm{\alpha}^\prime)$ is
  \begin{equation*}
    p_{\bm{\alpha}^\prime}(\bm{x}) = \frac{(m^\prime - 1)!}{(\sum_{j=1}^{m^\prime} \alpha^{\prime}_j x_j^{-1})^{m^\prime}}
    \prod_{k=1}^{m^\prime} \frac{\alpha^{\prime}_k}{x_k^2},
  \end{equation*}
  which is defined for all $x\in \Delta^{m^\prime - 1}$ with $x_j > 0$ for all $j\in [m^\prime]$.
  Furthermore, we have $p_{\bm{\alpha}^\prime}(\bm{x}) > 0$ for all $x\in \operatorname{int}\Delta^{m^\prime - 1}$
  since, in this case, $p_{\bm{\alpha}^\prime}(\bm{x})$ consists of a sum and products of a finite number of positive elements.
  By definition of $z \sim \operatorname{GS}(\bm{\alpha})$ we reverse the projection $P_{\bm{\alpha}}$ to obtain
  $\operatorname{supp}\bigl(\operatorname{GS}(\bm{\alpha})\bigr)=\operatorname{int}\{\bm{y}\in\mathbb{R}^n\;|\;y_j\in[0,1],\sum_{j=1}^m y_j = 1, y_k = 0 \text{ for } k\in J\}$.\\
  We will now show that $\operatorname{supp}(f)=(\frac{j_{\text{min}}}{m-1}, \tfrac{j_{\text{max}}}{m-1})$.
  First, let $\bm{z}\in\operatorname{supp}\bigl(\operatorname{GS}(\bm{\alpha})\bigr)$, then it holds that
  \begin{equation*}
    f(z) = \frac{1}{m-1} \sum_{j=1}^m j z_j = \frac{1}{m-1} \sum_{j=j_{\text{min}}}^m j z_j > \frac{1}{m-1} j_{\text{min}}.
  \end{equation*}
  With the same argument, we can show that $f(z) < \frac{j_{\text{max}}}{m-1}$.
  Conclusively, we will show that
  \begin{equation*}
    \forall \tilde{z} \in (\frac{j_{\text{min}}}{m-1}, \frac{j_{\text{max}}}{m-1})\; \exists \bm{z} \in \operatorname{supp}\bigl(\operatorname{GS}(\bm{\alpha})\bigr) \text{ with } \tilde{z} = f(\bm{z}).
  \end{equation*}
  Let $\tilde{z} \in (\frac{j_{\text{min}}}{m-1}, \frac{j_{\text{max}}}{m-1})$, then there exists $\delta \in (0,1)$ with
  $\tilde{z} = \delta \frac{j_{\text{min}}}{m-1} + (1-\delta) \frac{j_{\text{max}}}{m-1}$.
  Choose $\bm{z}$ with
  \begin{equation*}
    z_j =
    \begin{cases}
      \delta, & \text{if}\ j=j_{\text{min}}, \\
      1-\delta, & \text{if}\ j=j_{\text{max}}, \\
      0, & \text{otherwise}
    \end{cases}
  \end{equation*}
  to conclude the proof of Part (a).\\

  Part (b):
  Let $c>0$.
  We will first show that $\operatorname{GS}(\bm{\alpha}) = \operatorname{GS}(c\bm{\alpha})$.
  It holds that
  \begin{equation*}
    p_{c\bm{\alpha}}(\bm{x})
    = \frac{(m - 1)!}{(\sum_{j=1}^{m} c\alpha_j x_j^{-1})^{m}}
    \prod_{l=1}^{m} \frac{c\alpha_l}{x_l^2}
    = \frac{(m - 1)!}{(c\sum_{j=1}^{m} \alpha_j x_j^{-1})^{m}}
    c^m \prod_{l=1}^{m} \frac{\alpha_l}{x_l^2}
    % = \frac{c^m(m - 1)!}{c^m(\sum_{j=1}^{m} \alpha_j x_j^{-1})^{m}}
    % \prod_{l=1}^{m} \frac{\alpha_l}{x_l^2}
    = p_{\bm{\alpha}}(\bm{x}).
  \end{equation*}
  We will now show that $\frac{\alpha_k}{\alpha_j} \rightarrow 0$ for all $j\neq k$ and thus, $\frac{\bm{\alpha}}{\alpha_j}\rightarrow \bm{e}^j$ with
  \begin{equation*}
    e^j_k =
    \begin{cases}
      1, & \text{if}\ k=j, \\
      0, & \text{otherwise}
    \end{cases}
  \end{equation*}
  and therefore, $\operatorname{GS}(\bm{\alpha}) = \operatorname{GS}(\frac{1}{\alpha_j}\bm{\alpha}) \rightarrow \operatorname{GS}(\bm{e}^j)$ to conclude the proof.
  By assumption, we have
  \begin{equation*}
    \frac{1}{\sum_{k=1}^m \frac{\alpha_k}{\alpha_j}}
    = \frac{\frac{\alpha_j}{\alpha_j}}{\frac{1}{\alpha_j}\sum_{k=1}^m \alpha_k}
    = \frac{\alpha_j^{-1}}{\alpha_j^{-1}} \frac{\alpha_j}{\sum_{k=1}^m \alpha_k}
    = \frac{\alpha_j}{\sum_{k=1}^m \alpha_k}
    \rightarrow 1
  \end{equation*}
and thus, $\sum_{k=1}^m \frac{\alpha_k}{\alpha_j} \rightarrow 1$.
It holds that $\sum_{k=1}^m \frac{\alpha_k}{\alpha_j} = 1 + \sum_{j\neq k} \frac{\alpha_k}{\alpha_j}$ and thus, $\sum_{j\neq k} \frac{\alpha_k}{\alpha_j} \rightarrow 0$.
Since $\frac{\alpha_k}{\alpha_j} \geq 0$ for all $j\neq k$, we have $\frac{\alpha_k}{\alpha_j} \rightarrow 0$ and the proof follows.
\end{proof}

\subsubsection{Proof of Proposition~\ref{prop:rotation}}
\begin{proof}
    For the sake of clarity, we write $R \coloneqq R_{ij}^\alpha$.
    We know that $R\bm{z} \overset{d}{=} \bm{y}^{\prime}$ with
    $\bm{y}^{\prime}\sim\mathcal{N}\bigl(R\bm{\mu}, R\Sigma R^{\intercal}\bigr)$.
    Thus, we need to show that $R\Sigma R^{\intercal} = \Sigma$.
    Let $\hat{\sigma}\coloneqq\sigma_i=\sigma_j$.
    In the case of $n=2$, we have that
    $\bm{\sigma}=(\hat{\sigma}, \hat{\sigma})$ and
    \begin{equation*}
        R\Sigma R^{\intercal}
         = R\bm{\sigma}\bm{I}R^{\intercal}
         = R\hat{\sigma}\bm{I}R^{\intercal}
         = \hat{\sigma} R R^{\intercal}
         = \bm{\sigma}\bm{I}
         = \Sigma.
    \end{equation*}
    In the case of $n>2$, we use a change of basis to rotate over the first two axes.
    Let $P$ be the permutation matrix that swaps
    $\bm{e}_1 \leftrightarrow \bm{e}_i$ and $\bm{e}_2 \leftrightarrow \bm{e}_j$.
    Then $P = P^{-1} = P^{\intercal}$ and
    \begin{align*}
        R\Sigma R^{\intercal}
        &= P^{\intercal}PRP^{\intercal}P\Sigma P^{\intercal}PR^{\intercal}P^{\intercal}P\\
        &= P^{\intercal}R_P\Sigma_P R_P^{\intercal}P\\
        &= P^{\intercal} \left[
            \begin{array}{cc}
                R_{12} &0 \\
                % 0 & \bar{\bm{I}}
                0 & \bm{I}_{n-2}
            \end{array}\right]
            \left[
            \begin{array}{cc}
                \hat{\sigma}\bm{I}_2 &0 \\
                % 0 & \bar{\bm{\sigma}} \bar{\bm{I}}
                0 & \bar{\bm{\sigma}} \bm{I}_{n-2}
            \end{array}\right]
            \left[
                \begin{array}{cc}
                    R_{12}^{\intercal} &0 \\
                    % 0 & \bar{\bm{I}}
                    0 & \bm{I}_{n-2}
                \end{array}\right] P\\
        &= P^{\intercal} \left[
            \begin{array}{cc}
                R_{12}\hat{\sigma}\bm{I}_2 R_{12}^{\intercal} &0 \\
                % 0 & \bar{\bm{\sigma}} \bar{\bm{I}}
                0 & \bar{\bm{\sigma}} \bm{I}_{n-2}
            \end{array}\right] P\\
        &= P^{\intercal} \left[
            \begin{array}{cc}
                \hat{\sigma}\bm{I}_2 &0 \\
                % 0 & \bar{\bm{\sigma}} \bar{\bm{I}}
                0 & \bar{\bm{\sigma}} \bm{I}_{n-2}
            \end{array}\right] P\\
        &= P^{\intercal}\Sigma_P P\\
        &= \Sigma.
    \end{align*}
\end{proof}
%
%
%%%%%%%%%%%%%%%%%%%%%%%%%%%%%%%%
% Further theoretical considerations
%%%%%%%%%%%%%%%%%%%%%%%%%%%%%%%%
%
\section{Further theoretical considerations} \label{app:subs:neighX}
Without any inductive biases, unsupervised disentanglement is theoretically impossible \cite{locatello2019challenging}.
Fortunately, the negative ELBO loss function from Eq.~\ref{eq:elbo} imposes an inductive prior that encourages disentanglement \cite{burgess2018understanding}.
In this subsection, we discuss the concept of defining neighborhoods in the observable space.
We hypothesize that the closest neighbors of an observation typically differ in only a single dimension of the ground-truth factors.
\subsubsection{Defining neighborhoods in the observable space.}
Locatello et al.~\cite{locatello2019challenging} showed that there is an infinite number of transformations of the ground truth factors $\bm{z}\sim p(\bm{z})=\prod{p(z_i)}$ that lead to the same data distribution.
A representation $r(\bm{x})$ that is fully disentangled with respect to $\bm{z}$ might be fully entangled with respect to such a transformation $\hat{\bm{z}}$.
Without any inductive biases, unsupervised disentanglement is theoretically impossible.
We will make use of two properties to mitigate this impossibility result.
First, we can utilize the reconstruction loss to define \emph{neighboring observations}
\begin{equation}
  U_{\epsilon}(\bm{x}) = \left\{\bm{y}\;|\;-\E{q_{\phi}(\bm{z}|\bm{x})}{\log p_{\theta}(\bm{y}|\bm{z})} \leq \log\epsilon\right\}.  
\end{equation}
Intuitively, the neighborhood $U_{\epsilon}(\bm{x})$ of some observation $\bm{x}$ are those observations/reconstructions $\bm{y}$ that have a high log-likelihood when encoding $\bm{x}$.
This intuition becomes especially clear in the case of the mean squared error reconstruction loss since this loss function fulfills the properties of a metric.
In this case, the neighborhood simplifies to $U_{\epsilon}(\bm{x}) = \left\{\bm{y}\;|\;\frac{1}{d}\Vert \bm{x}-\bm{y} \Vert_2^2 \leq \epsilon\right\}$, and neighboring observations are those with similar pixel values.
We utilize a second property to associate neighboring observations with small changes in the ground truth factors.
Many datasets in the disentanglement literature consist of \emph{discrete} ground truth factors
\cite{lecun2004learning,reed2015deep,higgins2017beta,kim2018disentangling,locatello2019challenging,gondal2019transfer}.
We argue that because of the discrete nature of many datasets, e.g., pixels, even continuous ground truth factors often convert into discrete changes in the data space.
For instance, although we sample the X-position in the Circles dataset \cite{watters2019spatial} from a random uniform distribution, we only obtain $\sim40$ distinct observations regarding the X-position, see Figure~\ref{fig:circles} (left).
As a consequence, we mostly observe incremental changes in the ground truth factors $\bm{z}$, that is, a small change in a single dimension $z_i$ or $z_j$ or both, but never \emph{half} a change in $z_i$ and $z_j$ as illustrated in Figure~\ref{fig:latent_distance} (left).
We hypothesize that, consequently, the closest neighbors $\bm{x}^\prime$ of $\bm{x}$ are generally those observations whose ground-truth factors $\bm{z}^\prime$ differ in only a \emph{single} dimension compared to the ground-truth factor $\bm{z}$ of $\bm{x}$.
In the following, we will discuss neighborhoods in the latent space and eventually show that neighboring points in the data space are encouraged to be represented close together in the latent space enabling disentangling properties.
%
%
%
%%%%%%%%%%%%%%%%%%%%%%%%%%%%%%%%%%%%%%%%%%%%%%%%%%%%%%%%%%%%%%%%
%%%%%%%%%%%%%%%%%%%%%%%%%%%%%%%%%%%%%%%%%%%%%%%%%%%%%%%%%%%%%%%%
% Improving discrete representations
%%%%%%%%%%%%%%%%%%%%%%%%%%%%%%%%%%%%%%%%%%%%%%%%%%%%%%%%%%%%%%%%
%%%%%%%%%%%%%%%%%%%%%%%%%%%%%%%%%%%%%%%%%%%%%%%%%%%%%%%%%%%%%%%%
%
\section{Improving Discrete Representations} \label{app:improv}
Regularization and supervision encouraging disentangling properties play little to no role in models based on discrete latent spaces.
In this section, we demonstrate how to utilize some of the main results from the disentanglement literature to further improve the discrete representations of categorical VAEs.
%
%
%
%%%%%%%%%%%%%%%%%%%%%%%%%%%%%%%%
% Total correlation
%%%%%%%%%%%%%%%%%%%%%%%%%%%%%%%%
%
\subsubsection{Regularizing the total correlation.}
State-of-the-art unsupervised disentanglement methods enrich the Gaussian ELBO with various regularizers encouraging disentangling properties.
Kim \& Mnih \cite{kim2018disentangling} and Chen et al.~\cite{chen2018isolating} penalize the \emph{total correlation} \cite{watanabe1960information}
\begin{equation*}
  \operatorname{TC}(\bm{z}) =
  \kl{q(\bm{z})}{\hat{q}(\bm{z})} =
  \E{q(\bm{z})}{\log\frac{q(\bm{z})}{\hat{q}(\bm{z})}}
\end{equation*}
where $\hat{q}(\bm{z}) \coloneqq \prod_{i=1}^n q(z_i)$ to reduce the dependencies between the dimensions of the representation.
Kim \& Mnih \cite{kim2018disentangling} first sample from $\hat{q}(\bm{z})$ by randomly shuffling samples from $q(\bm{z})$ across the batch for each latent dimension \cite{arcones1992bootstrap}.
They then utilize the density-ratio trick \cite{nguyen2010estimating,sugiyama2012density} to estimate the total correlation by training a discriminator $D$ to classify between samples from $q(\bm{z})$ and $\hat{q}(\bm{z})$.
Fortunately, we can adopt the same procedure to estimate the total correlation of $q(\bar{\bm{z}})$ of the D-VAE latent variable.
We augment the ELBO of the D-VAE with a total correlation regularizer to obtain the learning objective
\begin{equation}\label{eq:factor_dvae}
    \mathcal{L}_{\theta,\phi}(\bm{x}) - \gamma \E{q(\bm{z})}{\log\frac{D(\bm{\bar{z}})}{1 - D(\bm{\bar{z}})}}
\end{equation}
for $\gamma>0$ and name the corresponding model \emph{FactorDVAE}.
Finding new regularizers of the total correlation, which are tailored to the D-VAE could be interesting future work.
%
%
%
%%%%%%%%%%%%%%%%%%%%%%%%%%%%%%%%
% Semi-supervised training
%%%%%%%%%%%%%%%%%%%%%%%%%%%%%%%%
%
\subsubsection{Semi-supervised training.}
The idea of semi-supervised disentanglement is that incorporating label information of a limited amount of annotated data points during training encourages a latent space with desirable structure w.r.t. the ground-truth factors of variation \cite{locatello2019disentangling}.
The supervision is incorporated by enriching the ELBO with a regularizer $R_s(r(\bm{x}), \bm{z})$, where $R_s$ is a function of the annotated observation-label pairs.
Locatello et al.~\cite{locatello2019disentangling} normalize the targets $z_i$ to $[0,1]$ and propose the binary cross-entropy loss (BCE) or the $L_2$ loss for $R_s$.
In contrast, we discretize $\bm{z}$ by binning each dimension $z_i$ into $m$ bins and utilize the cross-entropy loss for $R_s$ obtaining the learning objective
\begin{equation}\label{eq:semi-sup}
    \mathcal{L}_{\theta,\phi}(\bm{x}) + \omega \sum_{i=1}^n z_i^j \log \frac{\alpha_i^j}{\sum_{k=1}^m \alpha_i^k}
\end{equation}
where $\omega>0$ and $z_i^j = 1$ if $z_i$ is in bin $j$ and $z_i^j = 0$ otherwise.

In order to utilize semi-supervised training, a set of data points needs to be annotated beforehand.
Different ground-truth factors of variation usually have a specific finite number of unique values they can take on, see Table~\ref{tb:data} in Appendix~\ref{app:data}.
It is unclear how to incorporate the knowledge about the number of unique variations in the Gaussian VAE.
Thus, previous work dismisses this information entirely \cite{locatello2019disentangling}.
In contrast, it is straightforward to implement this information in the D-VAE using masked attention as introduced for the transformer architecture \cite{vaswani2017attention}.
If we know that factor $z_i$ can assume a total of $m^\prime < m$ distinct values, we set the set of the $m^\prime$ active categories to be
$J_i = \{1 + \lfloor j \tfrac{m-1}{m^\prime-1} \rceil\}_{j=0}^{m^\prime-1} \subseteq [m]$ and set $\alpha_i^j = 0$ for all $j \not\in J_i$.
We experiment with both the masked and the unmasked semi-supervision.
%
%
%
%%%%%%%%%%%%%%%%%%%%%%%%%%%%%%%%
% Further experiments
%%%%%%%%%%%%%%%%%%%%%%%%%%%%%%%%
%
\section{Further experiments} \label{app:further}
We explore the usefulness of different disentanglement metrics for downstream tasks, revealing that the MIG score is the most reliable indicator of sample efficiency across different datasets.
\subsubsection{Which disentanglement metric is useful for downstream tasks regarding the sample complexity of learning?}
\begin{figure}[!hb]
\begin{center}
\centerline{
    \includegraphics[width=0.6\columnwidth]{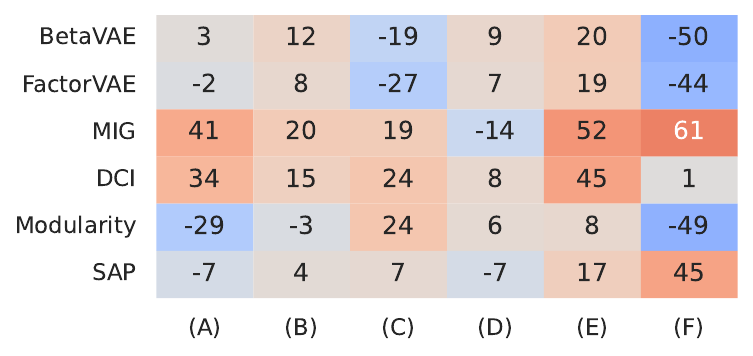}
}
\caption{The statistical efficiency of the simple downstream classification task of recovering the true factors of variations from the learned representation using gradient boosted trees (GBT).
A high MIG score reliably leads to a higher sample efficiency for all datasets but Cars3D.
The DCI score yields a positive correlation with the statistical efficiency.}
\label{fig:app_eff_gbt}
\end{center}
\end{figure}

In this experiment, we want to determine which disentanglement metric indicates a sound discrete latent space with respect to downstream tasks.
We follow the simple downstream classification task from \cite{locatello2019challenging} of recovering the true factors of variations from the learned representation using either multi-class logistic regression (LR) or gradient-boosted trees (GBT).
More precisely, we sample training sets of two different sizes, $100$ and $10\;000$, and evaluate the average test accuracy across factors on a test set of size $5\;000$, respectively.
To analyze the sample complexity, we measure the Spearman rank correlation between the different disentanglement metrics and the statistical efficiency that is, the test accuracy based on $100$ training samples divided by the accuracy based on $10\;000$ samples.
The right side of Figure~\ref{fig:st-gap_downstream} depicts this correlation regarding the LR task for all six datasets.
We can observe a high variance of the correlation depending on the selected disentanglement metric.
The correlation with the DCI, Modularity, and SAP scores depends on the data, while a high BetaVAE or FactorVAE score even negatively impacts the statistical efficiency.
Only a high MIG score reliably leads to a higher sample efficiency over all six datasets.
The experiments regarding the GBT task in Figure~\ref{fig:app_eff_gbt} mostly confirm this finding.
Consequently, we are mainly interested in the structural behavior of discrete representations regarding the MIG disentanglement score.
%
%
%
%%%%%%%%%%%%%%%%%%%%%%%%%%%%%%%%
% Implementation Details
%%%%%%%%%%%%%%%%%%%%%%%%%%%%%%%%
%
\section{Implementation details} \label{app:impl}
Locatello et al. \cite{locatello2019challenging} unified the choice of architecture, batch size, and optimizer to guarantee a fair comparison among the different methods.
We adopt these unifications and describe them here for the sake of completeness.
The only differences emerge from the Gumbel-softmax distribution from Equation~\ref{eq:gum_softmax}.
For all experiments, we choose the same number of $m=64$ categories.
If not mentioned differently, we utilize the symmetric interval $[-1,1]$ for the latent variable.
As proposed in \cite{friede2021efficient}, we utilize a constant Gumbel-softmax temperature of $\lambda=1.0$ and, instead, increase the scale parameter of the Gumbel distribution from $0.5$ to $2.0$ w.r.t. a cosine annealing and set the scale parameter to $0.0$ at test time.
We found this annealing scheme to improve training stability while encouraging discrete representations.
The implementation of the architectures is depicted in Table~\ref{tb:arch}, all hyperparameters can be found in Table~\ref{tb:hyper}.
We utilize the spatial broadcast decoder \cite{watters2019spatial} for the Circles experiments with a latent space dimension of $n=2$.
The implementations for the Circles experiments can be found in Table~\ref{tb:circle}.
If not mentioned differently, we utilize the ReLU activation function.
\begin{table*}[ht]
\caption{%
The architectures of the encoders and the decoder for the main experiments.}
\label{tb:arch}
\begin{center}
\begin{tabular}{l@{\hskip .3in}l@{\hskip .3in}l}
\toprule
\textbf{Encoder} (Gaussian) & \textbf{Encoder} (Discrete)   & \textbf{Decoder}\\
\midrule
Input: $64\times64\times C$ & Input: $64\times64\times C$   & Input: $10$ \\
Conv($4\times4,\;32,\;s=2$) & Conv($4\times4,\;32,\;s=2$)   & FC($256$) \\
Conv($4\times4,\;32,\;s=2$) & Conv($4\times4,\;32,\;s=2$)   & FC($4\times4\times64$) \\
Conv($4\times4,\;64,\;s=2$) & Conv($4\times4,\;64,\;s=2$)   & DeConv($4\times4,\;64,\;s=2$) \\
Conv($4\times4,\;64,\;s=2$) & Conv($4\times4,\;64,\;s=2$)   & DeConv($4\times4,\;32,\;s=2$) \\
FC($256$)                   & FC($256$)                     & DeConv($4\times4,\;32,\;s=2$) \\
FC($2\times10$)             & FC($10\times64$)              & DeConv($4\times4,\;C,\;s=2$) \\
\bottomrule
\end{tabular}
\end{center}
\end{table*}
\begin{table*}[ht]
\caption{%
The architectures of the discriminator for the TC regularizing experiments and the spatial broadcast decoder \cite{watters2019spatial} for the Circles experiments.}
\label{tb:circle}
\begin{center}
\begin{tabular}{l@{\hskip .5in}l}
\toprule
\textbf{Discriminator} & \textbf{Decoder} (Circles) \\
\midrule
FC($1000$), leaky ReLU      & Input: $2$    \\
FC($1000$), leaky ReLU      & Tile($64\times64\times10$)    \\
FC($1000$), leaky ReLU      & Concat. coordinate channels    \\
FC($1000$), leaky ReLU      & Conv($4\times4,\;64,\;s=1$)    \\
FC($1000$), leaky ReLU      & Conv($4\times4,\;64,\;s=1$)    \\
FC($1000$), leaky ReLU      & Conv($4\times4,\;C,\;s=1$)    \\
FC($2$)                     & \\
\bottomrule
\end{tabular}
\end{center}
\end{table*}
\begin{table*}[ht]
\caption{%
The model's hyperparameters.}
\label{tb:hyper}
\begin{center}
\begin{tabular}{l@{\hskip .3in}l@{\hskip .3in}l}
\toprule
\textbf{Parameter}              & \textbf{Model}            & \textbf{Values}\\
\midrule
Decoder type                &                   & Bernoulli  \\
Batch size                  &                   & $64$ \\
Latent space dim.      &                   & $10$ \\
Optimizer                   &                   & Adam \\
Adam: $\beta_1$             &                   & $0.9$ \\
Adam: $\beta_2$             &                   & $0.999$ \\
% Adam: $\epsilon$             &                   & $1e^{-8}$ \\
Learning rate               &                   & $1e^{-4}$ \\
Training steps              &                   & $300\;000$ \\
Latent space dim. (Circles) & Circles                  & $2$ \\
Number of categories        & discrete          & $64$ \\
Gumbel scale: init          & discrete          & $0.5$ \\
Gumbel scale: final         & discrete          & $2.0$ \\
Disc. Adam: $\beta_1$       & TC regularizing   & $0.5$ \\
Disc. Adam: $\beta_2$       & TC regularizing   & $0.9$ \\
% Disc. Adam: $\epsilon$             & TC regularizing                  & $1e^{-8}$ \\
$\gamma$                    & TC regularizing   & $[10,20,30,40,50,100]$ \\
$\omega$                    & semi-supervised   & $[1,2,4,6,8,16]$ \\
\bottomrule
\end{tabular}
\end{center}
\end{table*}
\begin{table*}[t]
\caption{%
The $25\%$ and the $75\%$ quantile MIG scores in \% for state-of-the-art unsupervised methods compared to the discrete methods.
Results taken from \cite{locatello2019challenging} are marked with an asterisk~(*).
We have re-implemented all other results with the same architecture as in \cite{locatello2019challenging} for the sake of fairness.}
\label{tb:mig_25}
\begin{center}
\begin{tabular}{lcccccc}
\toprule
Model & dSprites & C-dSprites & SmallNORB & Cars3D & Shapes3D & MPI3D\\
\midrule
$\beta$-VAE \cite{higgins2017beta}    &\scriptsize{[7.5,15.8]}$^*$&\scriptsize{[9.7,14.6]}$^*$& \scriptsize{[19.1,22.8]}$^*$&\scriptsize{[5.6,11.7]}$^*$& n.a.& n.a.\\
$\beta$-TCVAE \cite{chen2018isolating}  &\scriptsize{[13.6,22.2]}$^*$&\scriptsize{[10.4,18.0]}$^*$& \scriptsize{[18.3,24.5]}$^*$&\scriptsize{[7.3,14.0]}$^*$& n.a.& n.a.\\
DIP-VAE-I \cite{kumar2017variational}      &\scriptsize{[1.9,9.4]}$^*$&\scriptsize{[2.4,9.0]}$^*$& \scriptsize{[8.5,20.9]}$^*$&\scriptsize{[3.4,7.2]}$^*$& n.a.& n.a.         \\
DIP-VAE-II \cite{kumar2017variational}     &\scriptsize{[3.6,8.6]}$^*$&\scriptsize{[3.2,7.9]}$^*$& \scriptsize{[22.4,25.4]}$^*$&\scriptsize{[2.7,6.4]}$^*$& n.a.& n.a.\\
AnnealedVAE \cite{burgess2018understanding}    &\scriptsize{[2.9,20.9]}$^*$&\scriptsize{[4.8,25.7]}$^*$& \scriptsize{[1.5,8.1]}$^*$&\scriptsize{[4.6,7.7]}$^*$& n.a.& n.a.\\
FactorVAE \cite{kim2018disentangling}      &\scriptsize{[12.6,26.3]}&\scriptsize{[11.7,20.9]}& \scriptsize{[24.0,26.4]}&\scriptsize{[7.2,10.6]}&\scriptsize{[27.0,44.3]}&\scriptsize{[6.9,31.3]}\\
\midrule
D-VAE          &\scriptsize{[13.2,20.0]}&\scriptsize{[5.5,13.4]}&\scriptsize{[16.3,21.8]}& \scriptsize{[5.8,11.1]}& \scriptsize{[21.8,34.2]}&\scriptsize{[8.9,16.5]}\\
FactorDVAE    &\scriptsize{[14.5,35.7]}&\scriptsize{[11.3,20.3]}&\scriptsize{[20.6,24.8]}&\scriptsize{[12.8,16.3]}& \scriptsize{[34.8,48.3]}&\scriptsize{[26.0,32.1]}\\
\bottomrule
\end{tabular}
\end{center}
\end{table*}
%
%
%%%%%%%%%%%%%%%%%%%%%%%%%%%%%%%%
% Median Table
%%%%%%%%%%%%%%%%%%%%%%%%%%%%%%%%
%
\begin{table*}[ht]
\setlength{\tabcolsep}{3.5pt}
\caption{%
The $50\%$ (median), $25\%$, and $75\%$ quantiles in \% of the unsupervised D-VAE over all metrics.}
\label{tb:quant}
\begin{center}
\begin{tabular}{lcccccc}
\toprule
Metric & dSprites & C-dSprites & SmallNORB & Cars3D & Shapes3D & MPI3D\\
\midrule
BetaVAE&86.2&83.6&88.1&100.0&100.0&72.3\\
&\scriptsize{[85.4,86.6]}&\scriptsize{[81.9,85.0]}&\scriptsize{[85.6,90.2]}&\scriptsize{[100.0,100.0]}&\scriptsize{[99.5,100.0]}&\scriptsize{[67.9,76.5]}\\
FactorVAE&67.4&67.5&70.0&91.6&94.0&49.6\\
&\scriptsize{[61.9,71.9]}&\scriptsize{[60.3,71.3]}&\scriptsize{[66.9,73.3]}&\scriptsize{[89.0,94.0]}&\scriptsize{[88.3,98.2]}&\scriptsize{[46.7,53.4]}\\
MIG&17.4&9.4&19.0&8.5&28.8&12.8\\
&\scriptsize{[13.2,20.0]}&\scriptsize{[5.5,13.4]}&\scriptsize{[16.3,21.8]}&\scriptsize{[5.8,11.1]}&\scriptsize{[21.8,34.2]}&\scriptsize{[8.9,16.5]}\\
DCI&25.6&16.7&31.5&25.1&72.8&29.9\\
&\scriptsize{[19.6,28.0]}&\scriptsize{[12.7,20.4]}&\scriptsize{[29.5,32.7]}&\scriptsize{[21.0,29.4]}&\scriptsize{[68.1,78.2]}&\scriptsize{[27.6,31.7]}\\
Modularity&86.7&89.4&79.0&87.7&96.1&88.7\\
&\scriptsize{[84.5,88.6]}&\scriptsize{[87.0,91.2]}&\scriptsize{[76.5,81.3]}&\scriptsize{[85.8,89.3]}&\scriptsize{[94.9,97.2]}&\scriptsize{[87.2,89.9]}\\
SAP&6.6&2.5&8.6&1.4&7.4&5.5\\
&\scriptsize{[5.1,7.2]}&\scriptsize{[1.5,3.7]}&\scriptsize{[7.4,9.6]}&\scriptsize{[0.8,2.2]}&\scriptsize{[5.6,9.9]}&\scriptsize{[4.3,7.8]}\\
\bottomrule
\end{tabular}
\end{center}
\end{table*}
%
%
%
%
%
%
%%%%%%%%%%%%%%%%%%%%%%%%%%%%%%%%
% Median Table Semi-sup
%%%%%%%%%%%%%%%%%%%%%%%%%%%%%%%%
%
\begin{table*}[ht]
\setlength{\tabcolsep}{3.5pt}
\caption{%
The $50\%$ (median), $25\%$, and $75\%$ quantiles in \% of the MIG score for the discrete semi-supervised models D-VAE, D-VAE (Masked) (M), FactorDVAE, FactorDVAE (Masked) (M) for $1000$ labels.}
\label{tb:quant_semi}
\begin{center}
\begin{tabular}{lcccccc}
\toprule
Model & dSprites & C-dSprites & SmallNORB & Cars3D & Shapes3D & MPI3D\\
\midrule
D-VAE&32.0&28.4&15.8&17.7&45.8&39.2\\
&\scriptsize{[28.9,36.2]}&\scriptsize{[27.3,31.3]}&\scriptsize{[14.7,22.5]}&\scriptsize{[10.6,23.3]}&\scriptsize{[38.9,49.6]}&\scriptsize{[36.7,44.9]}\\
D-VAE (M)&32.0&27.2&25.3&22.0&46.0&52.1\\
&\scriptsize{[29.9,33.9]}&\scriptsize{[24.5,32.0]}&\scriptsize{[23.3,28.4]}&\scriptsize{[14.5,28.6]}&\scriptsize{[40.7,51.6]}&\scriptsize{[43.7,55.2]}\\
F-DVAE&37.6&37.4&27.2&11.4&38.8&36.5\\
&\scriptsize{[35.4,39.4]}&\scriptsize{[30.9,38.9]}&\scriptsize{[23.2,32.1]}&\scriptsize{[9.3,13.2]}&\scriptsize{[34.6,49.6]}&\scriptsize{[32.0,50.1]}\\
F-DVAE (M)&37.3&34.1&33.6&8.8&29.3&48.1\\
&\scriptsize{[29.6,38.4]}&\scriptsize{[23.2,37.0]}&\scriptsize{[26.8,37.3]}&\scriptsize{[5.9,10.1]}&\scriptsize{[23.0,42.3]}&\scriptsize{[35.2,52.7]}\\
\bottomrule
\end{tabular}
\end{center}
\end{table*}
%
%
%
%
%
%%%%%%%%%%%%%%%%%%%%%%%%%%%%%%%%
% Dataset Detail
%%%%%%%%%%%%%%%%%%%%%%%%%%%%%%%%
%
\section{Dataset details} \label{app:data}
All datasets are rendered in images of size $64\times64$ and normalized to $[0,1]$.
As in \cite{locatello2019challenging}, we directly sample from the generative model, effectively avoiding overfitting.
We consider gray-scale datasets dSprites, SmallNORB, and Circles, as well as datasets with three color channels C-dSprites, Cars3D, Shapes3D, and MPI3D.
We followed the instructions from \cite{watters2019spatial} to create the Circles dataset utilizing the Spriteworld environment \cite{watters2019spriteworld}, setting the size to $0.2$.
Table~\ref{tb:data} contains a set of all ground-truth factors of variation for each dataset.
\begin{table*}[ht!]
\caption{%
The ground-truth factors of the datasets.}
\label{tb:data}
\begin{center}
\begin{tabular}{l@{\hskip .3in}l@{\hskip .3in}c}
\toprule
\textbf{Dataset}            & \textbf{Ground-truth factor}        & \textbf{Number of values}\\
\midrule
dSprites               &  Shape                 & $3$  \\
                       &  Scale                 & $6$  \\
                       &  Orientation                 & $40$  \\
                       &  X-Position                 & $32$  \\
                       &  Y-Position                 & $32$  \\
\midrule
C-dSprites               &  Shape                 & $3$  \\
                       &  Scale                 & $6$  \\
                       &  Orientation                 & $40$  \\
                       &  X-Position                 & $32$  \\
                       &  Y-Position                 & $32$  \\
                       &  Color                 & Uniform$(0.5, 1.0)^3$  \\
\midrule
SmallNORB               &  Category                 & $5$  \\
                       &  Elevation               & $9$  \\
                       &  Azimuth                 & $18$  \\
                       &  Lighting condition                 & $6$  \\
\midrule
Cars3D                 &  Elevation               & $4$  \\
                       &  Azimuth                 & $24$  \\
                       &  Object type                 & $183$  \\
\midrule
Shapes3D               &  Floor color                 & $10$  \\
                       &  Wall color               & $10$  \\
                       &  Object color               & $10$  \\
                       &  Object size               & $8$  \\
                       &  Object type               & $4$  \\
                       &  Azimuth                 & $15$  \\
\midrule
MPI3D               &  Object color                 & $4$  \\
                       &  Object shape               & $4$  \\
                       &  Object size               & $2$  \\
                       &  Camera height               & $3$  \\
                       &  Background colors               & $3$  \\
                       &  First DOF                 & $40$  \\                       
                       &  Second DOF                 & $40$  \\
\midrule
Circles                & X-Position                 & Uniform($0.2, 0.8$)  \\
                       &  Y-Position                 & Uniform($0.2, 0.8$)  \\
\bottomrule
\end{tabular}
\end{center}
\end{table*}
%
%
%
%%%%%%%%%%%%%%%%%%%%%%%%%%%%%%%%
% Detailed experimental results
%%%%%%%%%%%%%%%%%%%%%%%%%%%%%%%%
%
\section{Detailed experimental results} \label{app:exp}
\subsubsection{Quantiles of the experimental results}
The $25\%$ and the $75\%$ quantile MIG scores in \% for state-of-the-art unsupervised methods compared to the discrete methods can be found in Table~\ref{tb:mig_25}.
The $50\%$ (median), $25\%$, and $75\%$ quantiles in \% of D-VAE over all metrics can be found in Table~\ref{tb:quant}.
The quantiles of the MIG score for the semi-supervised models with $1000$ labels can be found in Table~\ref{tb:quant_semi}.
\subsubsection{Circles experiment.}
\begin{figure}[t]
\begin{center}
\centerline{
    \includegraphics[width=0.9\columnwidth]{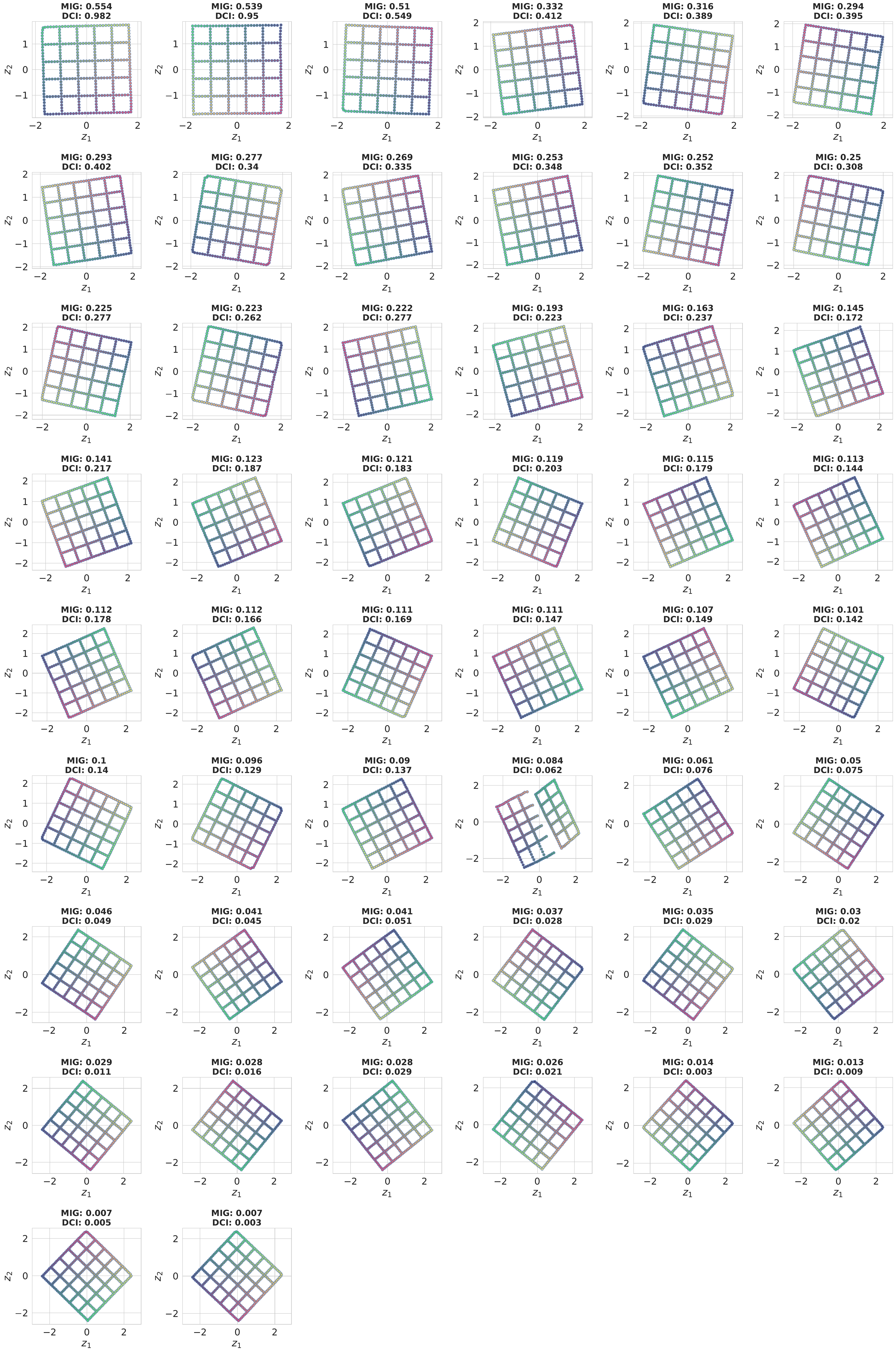}
}
\caption{A latent space geometry analysis of the circles experiment \cite{watters2019spatial} including the MIG and DCI scores.
We depict the latent space visualizations of all $50$ models of the Gaussian VAE sorted by the MIG score.}
\label{fig:50_vae}
\end{center}
\end{figure}

\begin{figure}[t]
\begin{center}
\centerline{
    \includegraphics[width=0.9\columnwidth]{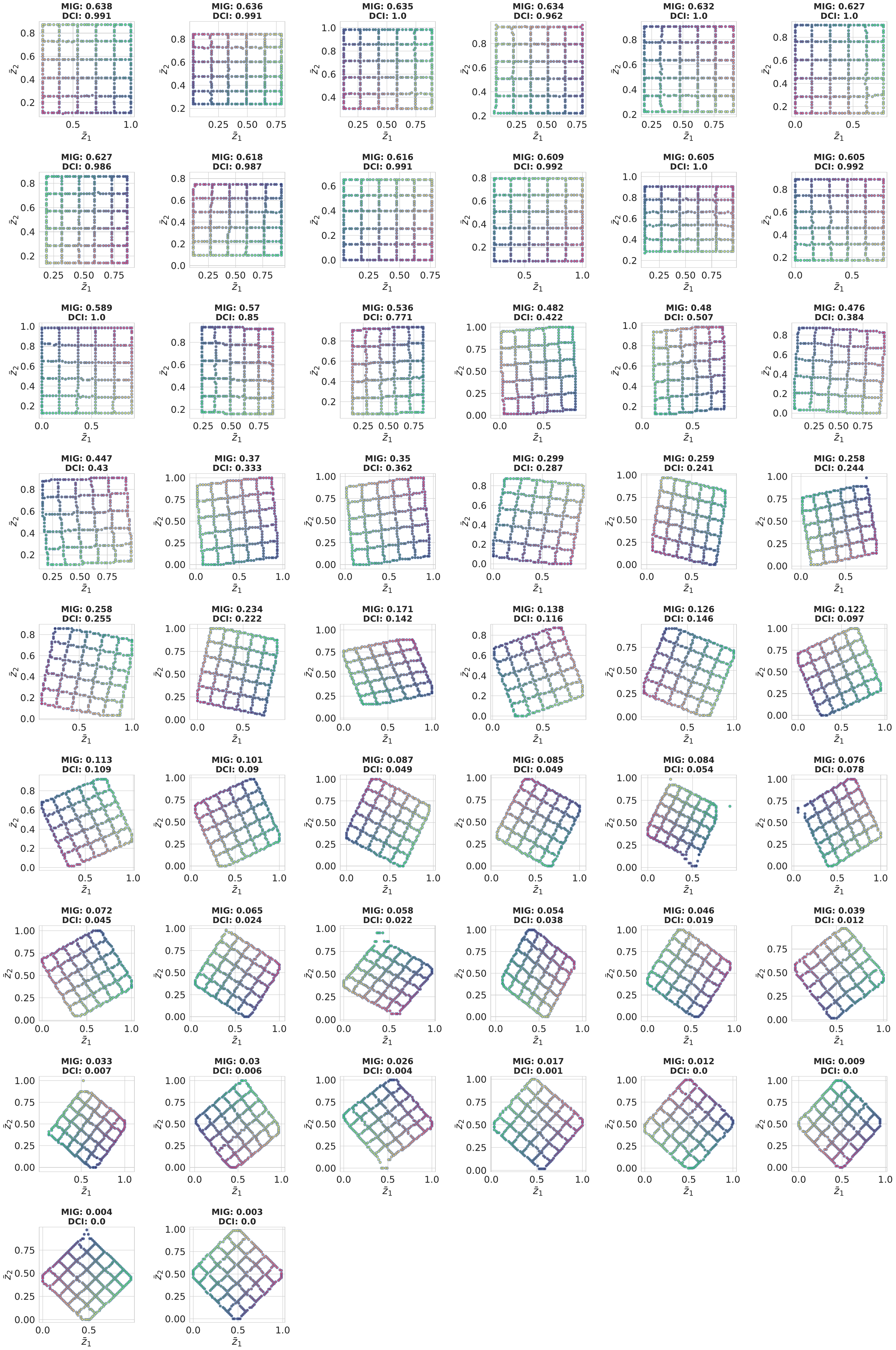}
}
\caption{A latent space geometry analysis of the circles experiment \cite{watters2019spatial} including the MIG and DCI scores.
We depict the latent space visualizations of all $50$ models of the discrete VAE sorted by the MIG score.}
\label{fig:50_d-vae}
\end{center}
\end{figure}

The latent space visualizations of the circles experiment \cite{watters2019spatial}, sorted by the MIG score of all $50$ models of the Gaussian VAE and the discrete VAE, respectively.
Figure~\ref{fig:50_vae} depicts the Gaussian latent spaces.
Even the latent spaces yielding the best MIG scores are affected by rotation.
Figure~\ref{fig:50_d-vae} depicts the discrete latent spaces.
More than $25$\% of the latent spaces lie parallel to the axes.
\subsubsection{Comparison of the unregularized models.}
\begin{figure}[t]
\begin{center}
\centerline{
    \includegraphics[width=0.9\columnwidth]{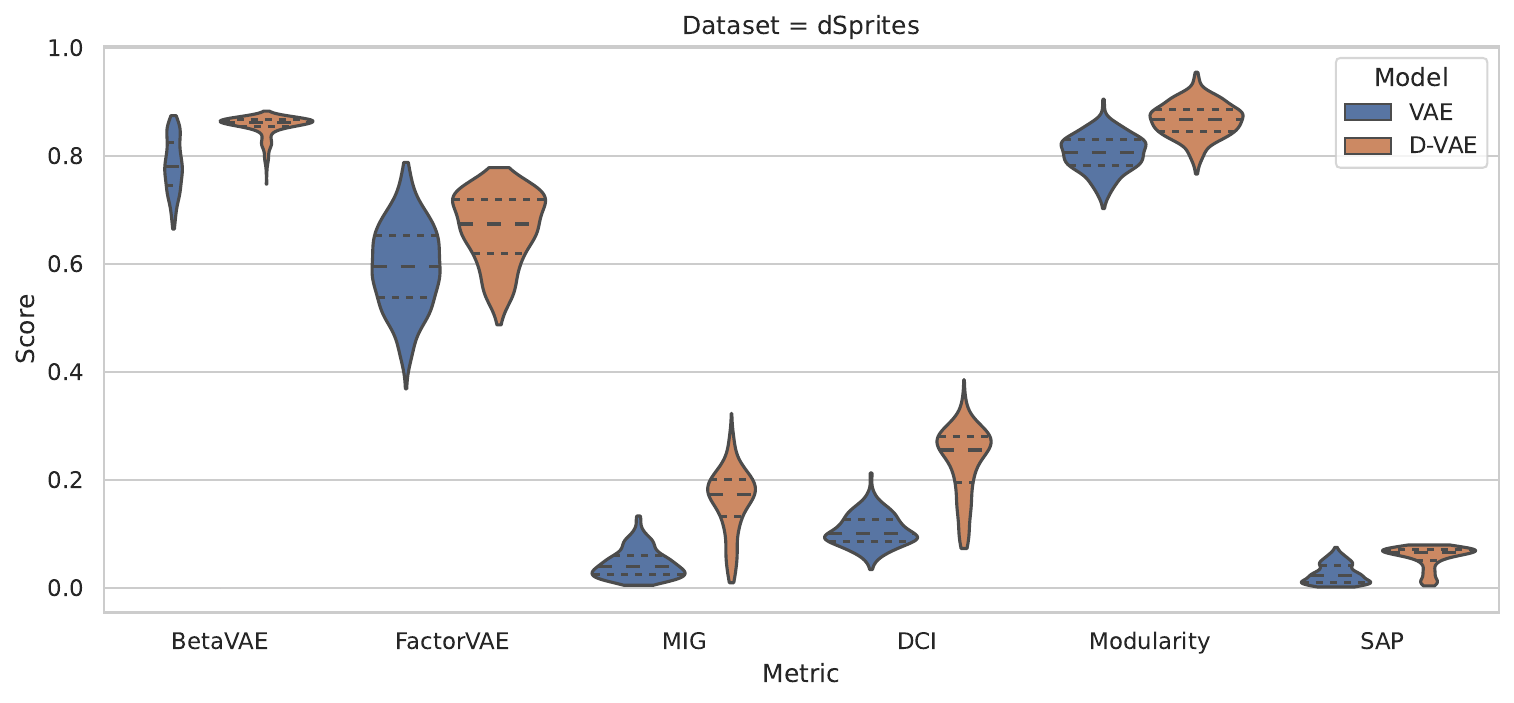}
}
\centerline{
    \includegraphics[width=0.9\columnwidth]{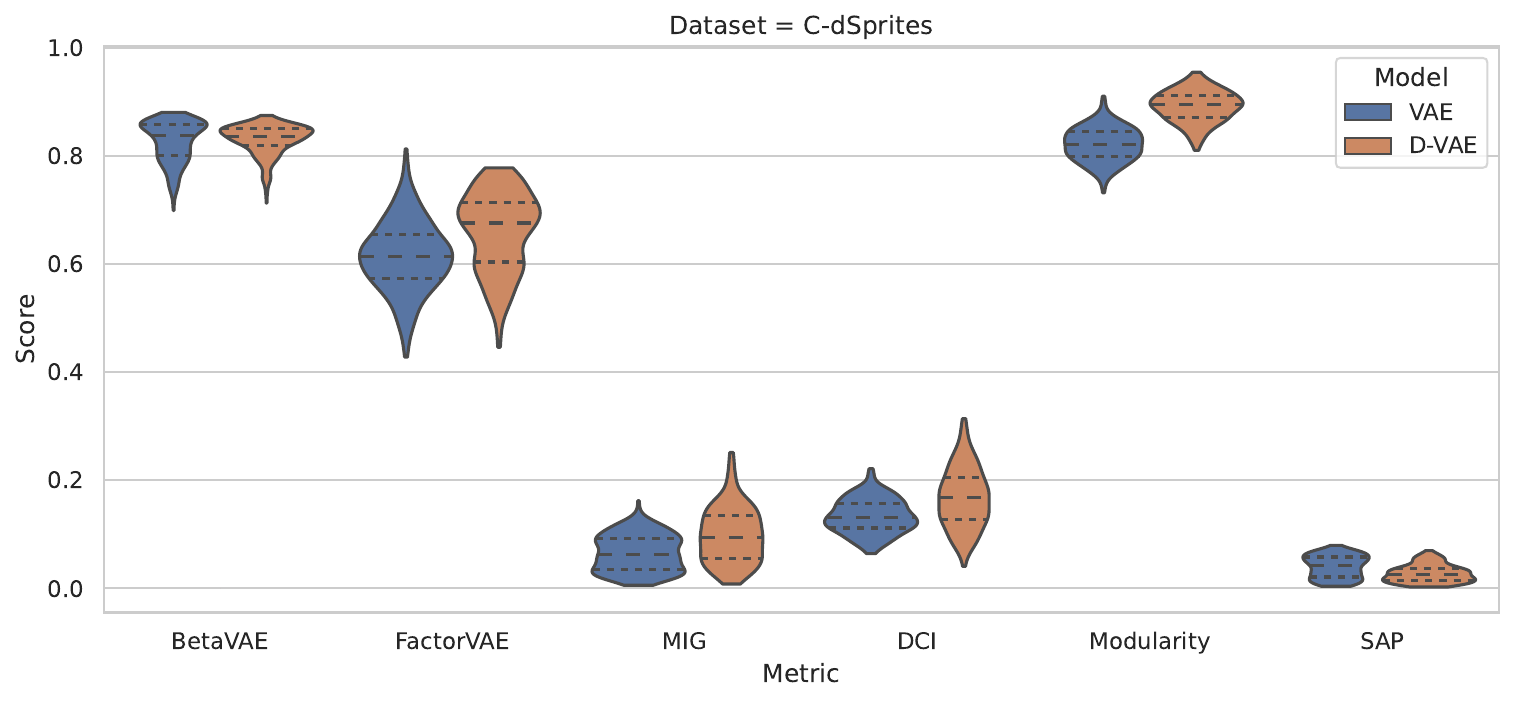}
}
\centerline{
    \includegraphics[width=0.9\columnwidth]{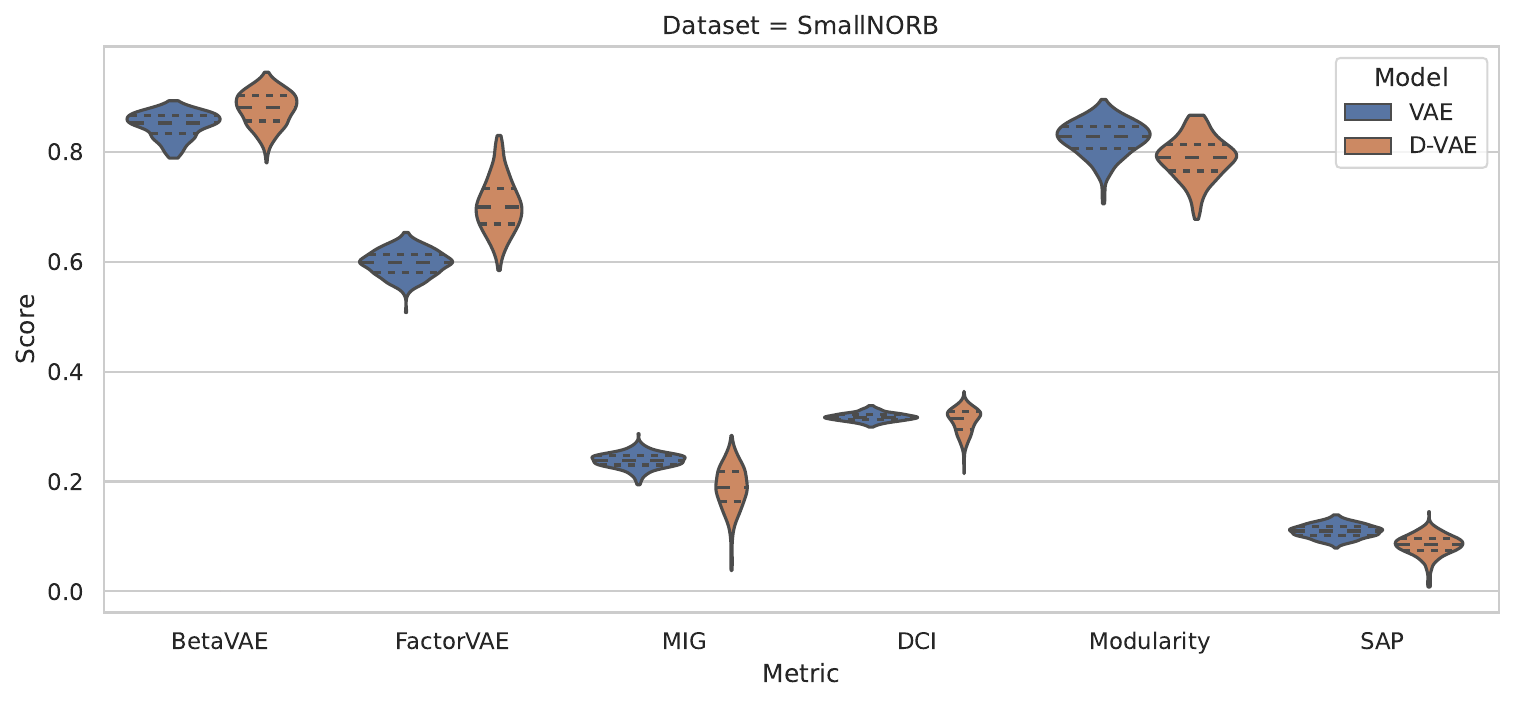}
}
\caption{A comparison of unregularized Gaussian VAE and the discrete VAE w.r.t. the 6 disentanglement metrics on dSprites, C-dSprites, SmallNORB.}
\label{fig:app_unreg_all_1}
\end{center}
\end{figure}

\begin{figure}[t]
\begin{center}
\centerline{
    \includegraphics[width=0.9\columnwidth]{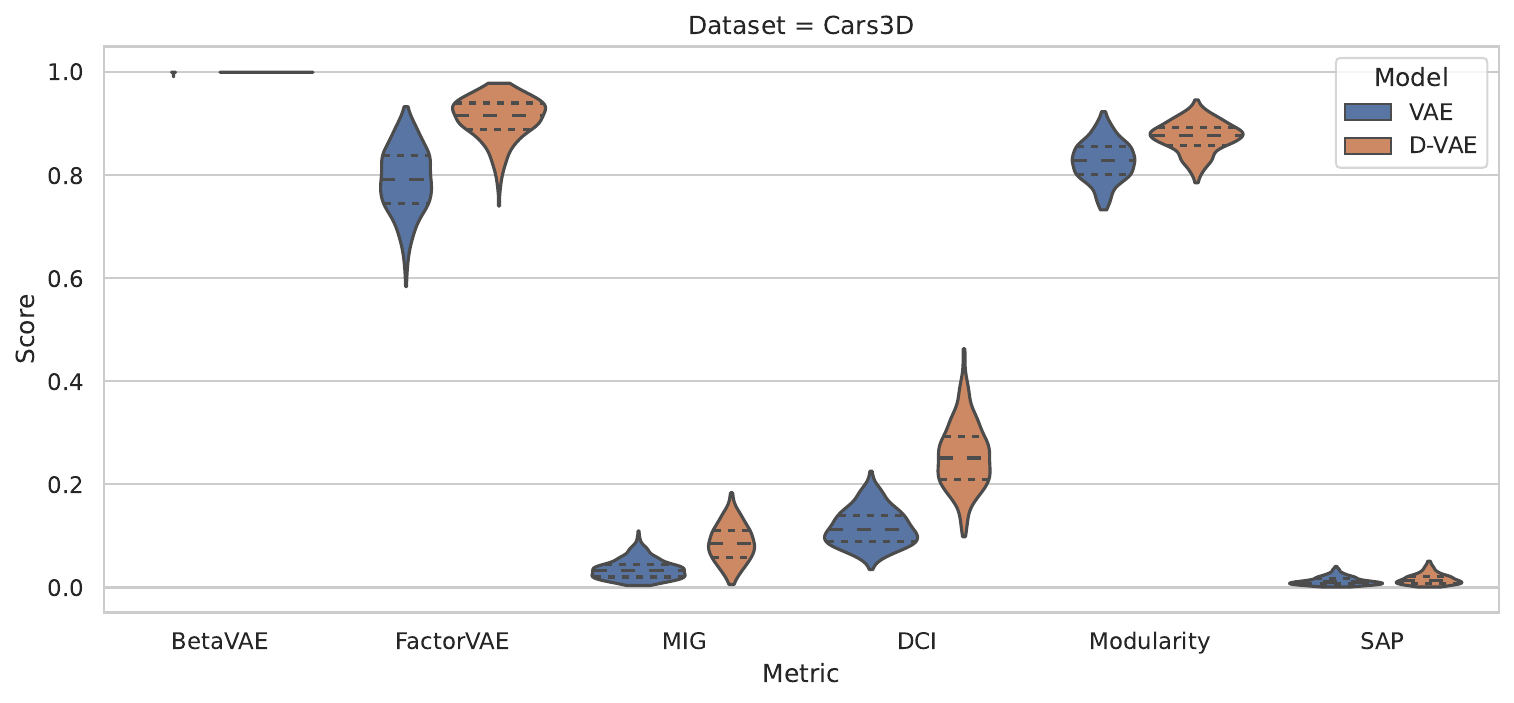}
}
\centerline{
    \includegraphics[width=0.9\columnwidth]{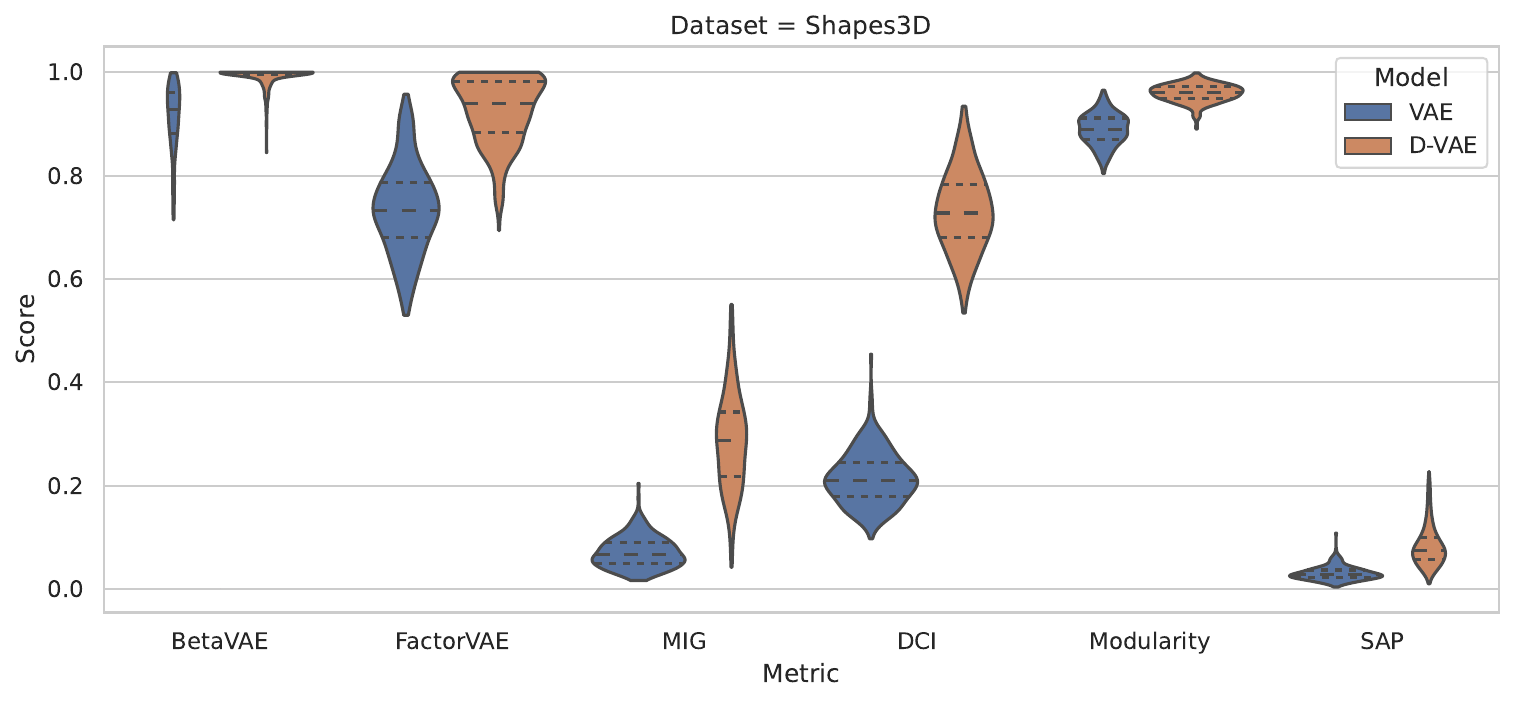}
}
\centerline{
    \includegraphics[width=0.9\columnwidth]{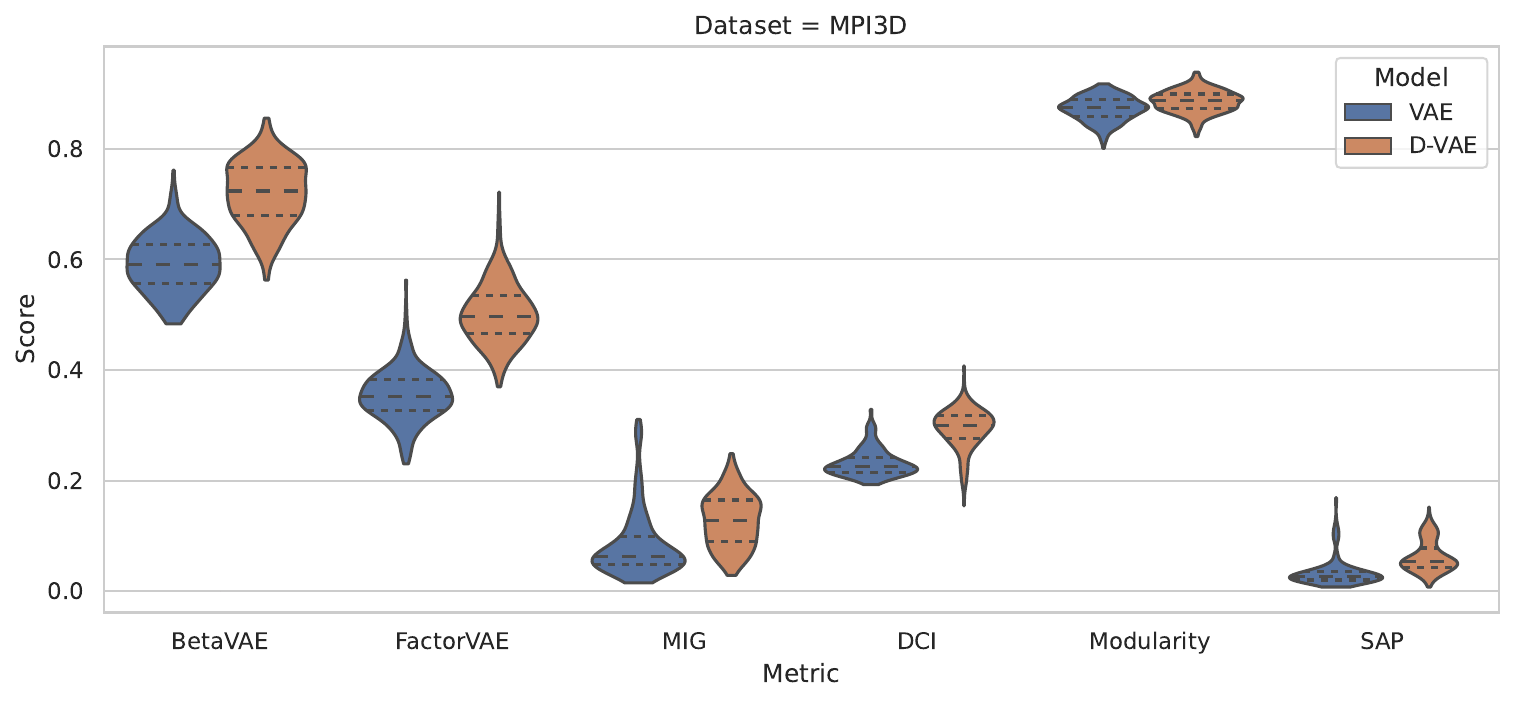}
}
\caption{A comparison of unregularized Gaussian VAE and the discrete VAE w.r.t. the 6 disentanglement metrics on Cars3D, Shapes3D, MPI3D.}
\label{fig:app_unreg_all_2}
\end{center}
\end{figure}

Figure~\ref{fig:app_unreg_all_1} and Figure~\ref{fig:app_unreg_all_2} depict the comparison of the unregularized models as violin plots for all datasets and metrics.
The discrete VAE improves over its Gaussian counterpart in $31$ out of $36$ cases.

\end{subappendices}